\documentclass[table]{article}
\usepackage{float}

\usepackage{PRIMEarxiv}

\usepackage[utf8]{inputenc} % allow utf-8 input
\usepackage[T1]{fontenc}    % use 8-bit T1 fonts
\usepackage{hyperref}       % hyperlinks
\usepackage{url}            % simple URL typesetting
\usepackage{booktabs}       % professional-quality tables
\usepackage{amsfonts}       % blackboard math symbols
\usepackage{nicefrac}       % compact symbols for 1/2, etc.
\usepackage{microtype}      % microtypography
      
\usepackage{lipsum}
\usepackage{fancyhdr}       % header
\usepackage{graphicx}       % graphics
\usepackage{float}
\graphicspath{{imgs/}}     % organize your images and other figures under media/ folder
\usepackage{amsmath,amssymb,amsfonts}
\usepackage{cite, textcomp}

\usepackage{multirow}
\usepackage{subcaption}
\usepackage[title]{appendix}%
\usepackage{graphicx}%
\usepackage{multirow}%
\usepackage{amsmath,amssymb,amsfonts}%
\usepackage{amsthm}%
\usepackage{mathrsfs}%
\usepackage[title]{appendix}%
 
\usepackage{textcomp}%
\usepackage{manyfoot}% 
\usepackage{algorithm}%
\usepackage{algorithmicx}%
\usepackage{algpseudocode}%
\usepackage{listings}%
\usepackage{tikz}
\usepackage{makecell}  
\usepackage{float}        
\usepackage{subcaption} 
\usepackage{orcidlink}
\usepackage{marvosym}  
\usepackage{orcidlink}
\usepackage{hyperref}
\usepackage{xspace}
\usepackage{subcaption}
 \usepackage{authblk}
\usepackage{array}         
\usetikzlibrary{positioning}
\usepackage{orcidlink}
\usepackage[justification=centering]{caption}
\usetikzlibrary{shapes, automata, positioning}

\usepackage{xcolor}     
\usepackage{colortbl}
\newcommand{\dataset}{\mathcal{D}} % Dataset
\newcommand{\Npatients}{\ensuremath{N}\xspace} %Total number of samples (patients) in the dataset.
\newcommand{\Nevents}{\ensuremath{K}\xspace} % Number of possible event types (e.g., different causes of failure in competing risks).
\newcommand{\Ncovariates}{\ensuremath{L}\xspace} % Number of covariates (features) per sample.

\newcommand{\eventi}{\ensuremath{y_{i}}} % indicador evento for patient i
\newcommand{\timei}{\ensuremath{t_{i}}} % time until the event for patient i
\newcommand{\covariablesi}{\ensuremath{x_{i}}} % vector covariables for patient i

\newcommand{\reparam}{\ensuremath{g}_{\phi}(\ensuremath{x_{i}}, \epsilon_i)}

\usepackage{geometry}
\title{Deep Survival Analysis in Multimodal Medical Data: A Parametric and Probabilistic Approach with Competing Risks}

\author[1]{Alba Garrido~\orcidlink{0009-0006-7238-1473}}
\author[1]{Alejandro Almodóvar~\orcidlink{0009-0006-0900-4026}}
\author[1]{Patricia A. Apellániz~\orcidlink{0000-0002-8604-9758}}
\author[1]{Juan Parras~\orcidlink{0000-0002-7028-3179}}
\author[1]{Santiago Zazo~\orcidlink{0000-0001-9073-7927}}

\affil[1]{Information Processing and Telecommunications Center, ETSI Telecomunicación, 

Universidad Politécnica de Madrid, Spain\\ 

\textit{Correspondence:} \href{mailto:alba.garrido.lopez@upm.es}{alba.garrido.lopez@upm.es}
}

\begin{document}
\maketitle
\thispagestyle{plain}

\begin{abstract}
\textbf{Purpose:} Accurate survival prediction is critical in oncology for prognosis and treatment planning. Traditional approaches often rely on a single data modality, limiting their ability to capture the complexity of tumor biology. To address this challenge, we introduce a multimodal deep learning framework for survival analysis capable of modeling both single and competing risks scenarios, evaluating the impact of integrating multiple medical data sources on survival predictions.

\textbf{Methods:} We propose SAMVAE (Survival Analysis Multimodal Variational Autoencoder), a novel deep learning architecture designed for survival prediction that integrates six data modalities: clinical variables, four molecular profiles, and histopathological images. SAMVAE leverages modality-specific encoders to project inputs into a shared latent space, enabling robust survival prediction while preserving modality-specific information. We evaluate SAMVAE on two cancer cohorts—breast cancer and lower-grade glioma—applying tailored preprocessing, dimensionality reduction, and hyperparameter optimization.

\textbf{Results:} The results demonstrate the successful integration of multimodal data for both standard survival analysis and competing risks scenarios across different datasets. Our model achieves competitive performance compared to state-of-the-art multimodal survival models. Notably, this is the first parametric multimodal deep learning architecture to incorporate competing risks while modeling continuous time to a specific event, using both tabular and image data.

\textbf{Conclusion:} SAMVAE introduces a probabilistic framework that supports the generation of personalized survival curves from heterogeneous data sources. Its parametric formulation enables the derivation of clinically meaningful statistics from the output distributions, providing patient-specific insights through interactive multimedia that contribute to more informed clinical decision-making and establish a foundation for interpretable, data-driven survival analysis in oncology.
\end{abstract}

% keywords can be removed

\section{Introduction}\label{sec1}
% General introduction
Healthcare data are inherently multimodal, involving diverse sources such as histopathological images, genomic sequencing, and structured clinical records \cite{bib1}. Traditional single-modality approaches, those relying on only one type of data, often fall short of capturing the complexity and richness of medical information \cite{bib5}. In contrast, multimodal approaches integrate and analyze multiple data types simultaneously, offering a more comprehensive view of patient health and disease mechanisms.

% Importance of multimodality 
The integration of multimodal data has become essential in oncology research \cite{bib2}. Furthermore, studies have increasingly highlighted the critical relationships between histopathological imaging and genetic events \cite{bib3}. In addition, the adoption of multi-modality approaches is a cornerstone of personalized medicine, as it enables an all-encompassing profile of each patient \cite{bib5}. This holistic perspective facilitates the development of tailored therapeutic strategies and more precise prognostic assessments, ultimately improving patient outcomes \cite{bib1}.

% The curse of dimensionality
Despite the promise of multimodal data in the advancement of cancer research, its high dimensionality presents both opportunities and challenges. Molecular and imaging datasets often involve high-dimensional and data-intensive formats; genomic profiles include thousands of features \cite{bib46}, while Whole Slide Images (WSIs) used in digital pathology often exceed dimensions of 100,000 × 100,000 pixels per slide \cite{bib45}. These modalities also differ greatly in how they are generated, pre-processed, and represented, making direct integration difficult due to discrepancies in scale, structure, and data format \cite{bib1, bib4}. This curse of dimensionality complicates the analysis and increases the risk of overfitting, particularly when the number of features far exceeds the number of samples. 

% Cancer, survival analysis and competing risks 
These challenges are especially critical in the context of medical research, where cancer remains one of the leading causes of death worldwide \cite{bib5}. A central objective in oncology is to estimate patient survival time, defined as the time until clinically significant events, such as disease progression or death, occur \cite{bib47}. To address this challenge, survival analysis provides statistical tools for modeling time-to-event data, including scenarios with competing risks, where alternative events may preclude or alter the occurrence of the primary event of interest \cite{bib48}.

% Limitations of SOTA
Despite numerous studies that address the problem of survival analysis,  many existing approaches face significant limitations. Some rely on oversimplified models that fail to capture the complexity and heterogeneity of cancer data \cite{bib5}. In contrast, multimodal models present additional limitations such as the lack of interpretability in their results \cite{bib15}, making their integration into clinical decision-making challenging. Furthermore, only a limited number of studies leverage truly multimodal data, and the lack of transparency in the code of some models, coupled with the inaccessibility of certain private datasets, significantly hinders reproducibility and fair benchmarking across the field, as reported by \textit{Wiegrebe et al.} \cite{bib15}.

% Architecture based on VAEs: interpretability
Building on these motivations, we propose SAMVAE (Survival Analysis Multimodal Variational Autoencoder), a multimodal integration framework for parametric survival analysis and competing risks. To address the inherent heterogeneity across different modalities of data, we adopt a flexible architecture based on Variational Autoencoders (VAEs) \cite{bib8}, generative models capable of capturing complex nonlinear relationships within data. This allows us to learn shared latent representations across modalities while enabling interpretability at the output level, as VAEs can model parametric distributions over both covariates and event times, provided the log-likelihood is differentiable \cite{bib7}. This parametric and fully probabilistic formulation is central to our approach, as it enables the principled estimation of uncertainty, allows for flexible distributional assumptions beyond proportional hazards \cite{bib50}, and provides analytical access to key survival quantities such as event probabilities and cumulative incidence functions.

To validate the proposed model, we conduct experiments on breast cancer and lower-grade glioma datasets, leveraging six distinct data modalities. In line with the general classification of oncology data into clinical, molecular, and imaging domains \cite{bib5}, our study incorporates clinical information, four molecular modalities (DNA methylation (DNAm), Copy Number Variation (CNV), microRNA (miRNA), and RNA sequencing (RNAseq)), and WSIs.  

 % General pipeline

 \begin{figure}[ht]
\centering
\includegraphics[width=0.8\textwidth]{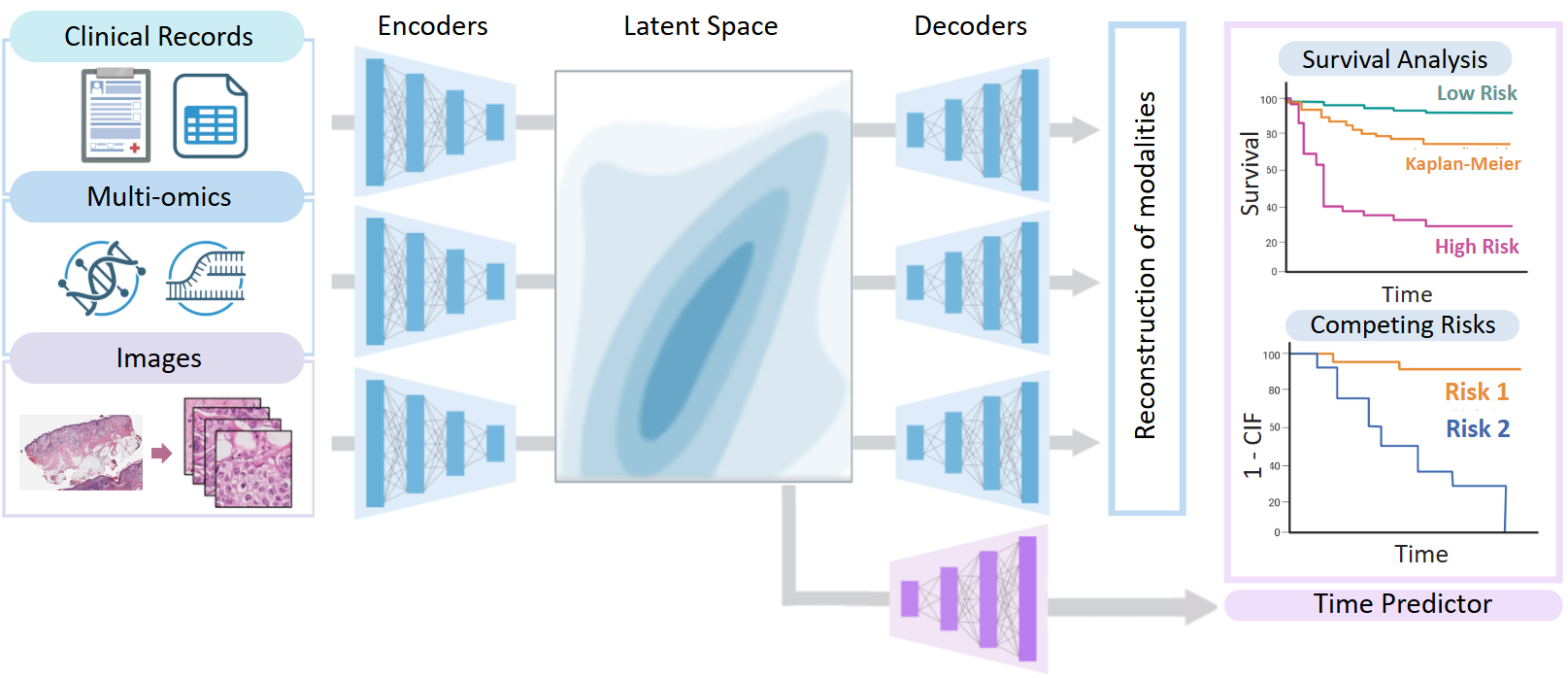}
\caption{General architecture of the proposed SAMVAE framework. The model receives multimodal input data, which are encoded into a shared latent space using modality-specific encoders. This latent representation is used to predict time-to-event outcomes, either survival analysis or competing risks modeling, depending on whether the dataset includes one or multiple mutually exclusive events. }\label{fig:samvaepipeline}
\end{figure}  

An overview of the proposed architecture is shown in Figure \ref{fig:samvaepipeline}, which outlines the core components of SAMVAE. The model employs modality-specific encoders to transform heterogeneous input data into a common latent space that captures integrated patient representations. This latent space is used for the time prediction, with the specific modeling strategy depending on the nature of the dataset: standard survival analysis is applied when there is a single event of interest, whereas competing risks modeling is used when multiple mutually exclusive events are present. This architecture supports flexible fusion and end-to-end training and optimization, enabling SAMVAE to effectively handle the complexity of multimodal biomedical data across different survival modeling scenarios.  
 
 Then, our main contributions consist of:  
\begin{itemize}
    \item We introduce SAMVAE, a novel multimodal extension of the SAVAE framework \cite{bib7}, designed to jointly model heterogeneous multimodal biomedical data. By employing modality-specific encoders and projecting them into a shared latent space, SAMVAE enables flexible integration of diverse data types with end-to-end trainability.
    \item We extend the SAMVAE framework to flexibly support both standard survival analysis and competing risks modeling, enabling task-specific adaptation within a unified multimodal architecture. To the best of our knowledge, this is the first multimodal deep learning framework to incorporate competing risks while modeling time-to-event outcomes in continuous time. This represents a significant step forward in personalized survival prediction across heterogeneous data sources (tabular and image data).
    \item We take advantage of the parametric and probabilistic formulation of SAMVAE to generate individualized, interpretable, and interactive survival curves. This facilitates a deeper understanding of the multimodal factors influencing patient-specific risk trajectories, thereby enhancing the model’s clinical transparency and potential for real-world applicability.
    \item We use publicly available datasets and release our full codebase, promoting transparency, reproducibility, and enabling independent validation.
    
\end{itemize}

\section{Background}\label{sec2}
\subsection{Survival Analysis}\label{sec2_1}
Survival Analysis, also known as time-to-event analysis, is a statistical method used to model the time until a specific event occurs. In healthcare, survival analysis plays a crucial role in predicting and understanding health-related events such as disease progression, hospital readmissions, and mortality \cite{bib51}. It supports tasks such as prognostic modeling, in which clinicians assess the likelihood of adverse outcomes for specific patient groups and identify high-risk individuals. A key feature of survival analysis is its ability to handle censored data—instances where the exact time of the event is unknown, either because the event has not yet occurred or the subjects have left a study \cite{bib6}.

We consider that we have a dataset with \Npatients  observations, each represented by triplets 
\( \dataset = \bigl(\covariablesi,\timei, \eventi \bigr)_{i=1}^\Npatients \). The vector \(\covariablesi = (\covariablesi^1, \dots, \covariablesi^\Ncovariates)\) comprises \(\Ncovariates\) covariates for individual \(i\), while \(\timei\) denotes the time at which an event occurs. The indicator \(\eventi \in \{0,1\}\) specifies whether the event was observed \((\eventi = 1)\) or censored \((\eventi = 0)\). When an observation is censored, it means that no event was recorded up to time \(\timei\) \cite{bib7}.

The survival function, \( S(t \mid x)\), is a non-increasing function of \( t \) that indicates the probability that the event of interest has not yet occurred by time \( t \), conditional on covariates \( x \). It is defined as \( S(t \mid x) = P(T > t \mid x) = 1 - F(t \mid x) \) where \( F(t \mid x) \) is the Cumulative Distribution Function (CDF) of event times.

\subsection{Competing Risks}\label{sec2_2}
 
Competing Risks refer to situations in survival analysis where multiple mutually exclusive event types can occur, and the occurrence of one event prevents the occurrence or observation of the others. The Cumulative Incidence Function (CIF) becomes the key quantity of interest in competing risks settings. 
Let $Y \in\{0,1,.., K\}$ be a discrete random variable, where $Y = 0$ denotes exclusively censoring, and $Y = k$ indicates the occurrence of event type $k$. The CIF is defined only for the actual event types $k \in\{ 1, 2,...,K\}$, and not for the censored case:

$$
\text{CIF}_k(t \mid x) = P(T \leq t, Y = k \mid x),
$$

where $T$ is the time-to-event, $k \neq 0$, and $Y = k$ indicates the event type. The competing risks model reduces to the standard survival analysis setting when $K=1$, that is, when $Y \in\{0,1\}$, meaning there can be either a single event of interest ($Y = 1$) or censoring ($Y = 0$). Unlike the survival function, which aggregates all failure types, the CIF separates cause-specific probabilities. Furthermore, the CIF does not converge to 1 as $t \to \infty$ and the sum across all risks equals the overall event probability, excluding censoring.

As shown by \textit{Apellániz et al.} \cite{bib9}, the CIF is defined as:
\begin{equation}
\text{CIF}_k(t \mid x) = P(T \leq t \mid Y = k, x) \cdot P(Y = k \mid x)
\label{eq:CIF}
\end{equation}

where the first term (the conditional survival probability for each event type) can be obtained via risk-specific predictors and the second (the risk probability) through a latent-variable classifier. For censored observations, the survival contribution is derived from the fact that no event occurred before the censoring time $t$, leading to the following definition:

\begin{equation}
  S(t \mid x) =    1 - \sum_{k=1}^{\Nevents} \text{CIF}_k(t\mid x) .
\end{equation}

\subsection{Related Works} 

 \subsubsection{Survival Analysis Models} 
Traditional survival analysis has been dominated by the Cox proportional hazards (CoxPH) model \cite{bib50}, which assumes a linear relationship between covariates and the log-risk function and constant hazard ratios over time. While CoxPH remains popular for its interpretability, its assumptions are often too restrictive for high-dimensional or time-varying data, leading to the development of several deep learning-based extensions. DeepSurv \cite{bib52} replaces the linear predictor in CoxPH with a deep neural network, enabling the capture of non-linear covariate interactions while maintaining the proportional hazards assumption. Extensions such as Cox-Time \cite{bib53} relax this assumption by allowing time-varying effects, while Cox-PASNet \cite{bib54} integrates prior biological knowledge and regularization for high-dimensional omics data.

Moving beyond the proportional hazards framework, models like DeepHit \cite{bib55} directly estimate the discrete-time survival distribution, allowing for flexible, non-linear, and non-monotonic risk modeling. Its extension, Dynamic-DeepHit \cite{bib56}, incorporates recurrent neural networks (RNNs) to handle longitudinal data with time-dependent covariates. Fully parametric deep survival models have also gained attention. Approaches such as Deep Weibull \cite{bib57} and Deep Survival Machines (DSM) \cite{bib58} assume parametric survival distributions and leverage mixture models to increase flexibility. Additionally, SAVAE \cite{bib7} builds on these parametric advances by introducing a VAE-based framework that models survival times through flexible parametric distributions. Its design enables effective handling of high-dimensional, heterogeneous, and censored data.

 \subsubsection{Competing Risks Models} 

The Fine-Gray (FG) model \cite{bib59} is one of the most established classical approaches for competing risks. By modeling the subdistribution hazard directly, it enables estimation of the CIF for each event type. However, like CoxPH, FG relies on the proportional hazards assumption and lacks flexibility in complex, real-world scenarios.

Deep learning has introduced more flexible alternatives. DeepHit \cite{bib55} estimates CIFs for multiple events using shared and cause-specific subnetworks, trained via a composite loss. Though powerful, it operates in discrete time and lacks closed-form CIFs. DeepComp \cite{bib60} builds on this by modeling cause-specific discrete hazards with RNNs, improving temporal modeling but still constrained by time discretization. Similarly, CRESA \cite{bib61} uses RNNs to model recurrent CIFs and is particularly suited for sequential data, though it shares the same discrete-time limitations.

To address these, SSMTL \cite{bib62} frames competing risks as a multi-task classification problem, enabling semi-supervised learning but reducing temporal granularity. In contrast, DeepCompete \cite{bib63} introduces a continuous-time framework based on neural ordinary differential equation blocks, offering more precise CIF estimation at the cost of increased computational complexity. CR-SAVAE \cite{bib9} proposes a generative VAE-based approach that enables accurate cumulative incidence estimation without relying on proportional hazards assumptions, valid for continuous and discrete time. It improves the interpretability of the results by allowing direct analysis of covariate effects and supports robust statistical inference, facilitating personalized and clinically meaningful risk assessment.

\subsubsection{State of the Art: Multimodal Survival Approaches in Medical Data}\label{sec2_3}

\paragraph{Survival Analysis Models} \label{sec2_31}  
\hfill 
\break 
  To assess the current state of the field of survival analysis using deep learning, a comprehensive survey \cite{bib15} published in 2024 initially identified 211 relevant articles in this domain, with an openly maintained codebase to support reproducibility. This review underscores a persistent gap: very few survival models fully exploit multimodal data using public repositories with transparent and standardized evaluation protocols.
 
During the literature review, several multimodal survival analysis models were identified.  Many potentially relevant studies relied on private or restricted datasets (e.g., hospital cohorts), limiting reproducibility and fair benchmarking \cite{bib27}, or lacked accessible code despite using public data, as in CapSurv \cite{bib29}. Therefore, many were excluded from direct comparison with our approach due to limitations in data types, target populations, or methodological scope. 

Our work focuses on integrating clinical data, histopathological images, and multi-omics in specific cancers.  The inclusion criteria required that each model incorporated three data modalities (clinical, omic, and images) and used The Cancer Genome Atlas (TCGA) public dataset \cite{bib10}, one of the most widely adopted and publicly available repositories. From these studies, only 29 addressed multimodality, and a subset of 6 studies utilized the widely recognized and publicly available TCGA repository and made their source code accessible on GitHub.

 This subset of six studies, although focused on multimodal integration in oncology, exhibits key limitations that distinguish them from our approach. AdvMIL \cite{bib32} focuses exclusively on histopathological images, without incorporating clinical or omics data. PAGE-Net \cite{bib33} integrates histological images with gene expression data but is tailored specifically for glioblastoma and relies on pathway-level genomic integration, diverging from our use of raw multi-omics data. GD-Net and ConcatAE/CrossAE \cite{bib30, bib35} incorporate only three omics modalities, omitting both clinical and imaging data, thereby falling short of full multimodal integration. MultiSurv \cite{bib31} supports multiple modalities but is trained across 33 cancer types simultaneously, lacking the tumor-specific optimization that is central to our study. Lastly, BioFusionNet \cite{bib34} employs a highly specialized and computationally intensive framework based on large-scale self-supervised pretraining on over 35 million histopathological patches (DINO and MoCoV3) and dual cross-attention mechanisms, processing up to 500 images per patient. This results in extremely high-dimensional embeddings from clinical, omics, and imaging data, accompanied by substantial computational costs.  The results reported for BioFusionNet and its comparative models were obtained by training on a single TCGA dataset, which integrates clinical, omics, and imaging modalities. Therefore, we compare our approach against the reported results of BioFusionNet in the literature and the other models referenced therein. 

In addition to BioFusionNet, several recent multimodal survival prediction models have been proposed but are not covered in the survey mentioned \cite{bib15}. MultiDeepCox-SC \cite{bib37} integrates image-derived risk scores with clinical and gene expression data using the Cox proportional hazards model \cite{bib50}. MCAT \cite{bib40} introduces a genomics-guided co-attention mechanism to align whole-slide images with genomic and clinical features. MultiSurv (mentioned above) \cite{bib31} concatenates feature vectors into a unified representation for survival estimation. HFBSurv \cite{bib38} leverages a hierarchical attention-based bilinear structure to process data progressively from simple to complex levels. PathomicFusion \cite{bib39} applies gating-based attention to suppress noise and highlight relevant features. TransSurv \cite{bib41} employs cross-attention transformers to combine histopathological, genomic, and clinical inputs, capturing intermodal dependencies. These approaches highlight the diversity of fusion strategies explored in multimodal survival modeling, with reported results on the TCGA breast cancer cohort available in the BioFusionNet benchmark, which we use as a basis for comparison in our experimental evaluation for survival analysis.

\paragraph{Competing Risks Models}\label{sec2_32}
\hfill 
\break 
To evaluate the current state of the field of competing risks using deep learning, the same comprehensive survey mentioned above was also reviewed \cite{bib15}, this time filtering specifically on methods designed to handle competing risks. A few relevant works were identified; however, they present notable limitations. For instance, DeepPAMM \cite{bib66} addresses competing risks and includes a multimodal experiment combining tabular data and point clouds. However, this model operates exclusively in the discrete-time domain because it uses a penalized negative Poisson log-likelihood as a loss function, which requires datasets with a large number of observations, an assumption that does not typically hold in real-world clinical datasets. Neural Fine-Gray \cite{bib67} focuses exclusively on tabular datasets and explicitly mentions the extension to other data types, such as time series or images, as future work. Therefore, we exclude it from comparison due to its current lack of support for multimodality. Lastly, SURVNODE \cite{bib68} explores multi-state models and competing risks, but its experiments are performed separately on either tabular or time series data, without integrating multiple modalities. These limitations underscore the need for models capable of jointly handling competing risks and multimodal inputs, such as tabular clinical or omic data and imaging—using real-world datasets.

In addition to the models covered in the survey, we identified an additional approach not included in it: Deep-CR MTLR \cite{bib69}, which includes both clinical and imaging modalities; however, it does not include an omics modality. In addition, this model operates exclusively in the discrete-time domain. This introduces several limitations, most notably, the need for a dataset with a large number of observations, a condition that is often unrealistic in real-world settings. Furthermore, training Deep-CR MTLR requires fitting a separate model for each time point and for each event type, making the approach computationally intensive.

To date, and to the best of our knowledge, no previous work has proposed a multimodal deep learning framework that explicitly models competing risks in continuous time. Existing multimodal approaches predominantly rely on conventional Cox proportional hazards models or focus on single-event survival settings, thereby overlooking the complexity introduced by mutually exclusive event types. In contrast, our study is the first, based on the available literature, to integrate competing risks modeling within a fully multimodal architecture encompassing clinical, imaging, and multi-omics data, with the flexibility to handle both continuous and discrete time-to-event data. Additionally, our approach is parametric and does not rely on the proportional hazards assumption. This versatility enhances
its applicability in real-world healthcare settings, where both data types and event structures are heterogeneous.

\section{Architecture: from Unimodal to Multimodal VAEs for Survival}\label{sec3}

In this section, we provide a detailed description of the SAVAE model \cite{bib7} and its extension to competing risks (CR-SAVAE \cite{bib9}), which serve as the foundation for our proposed SAMVAE. While these models were briefly introduced in the related work section, we now detail their structure and survival modeling framework in order to clarify the design choices behind our multimodal extension. Building on this foundation, we then introduce the architecture of SAMVAE, which generalizes the unimodal models to effectively integrate heterogeneous data sources.

\subsection{Unimodal SAVAE for Single and Competing Risks}

 SAVAE \cite{bib7} introduces a generative framework based on VAEs \cite{bib8}, allowing it to estimate survival times by parametrically modeling the underlying distribution whether Weibull, log-normal, or other differentiable forms. This removes the need for strong assumptions, such as proportional hazards, and enables the incorporation of complex, non-linear interactions between covariates and survival outcomes. Moreover, a tailored Evidence Lower BOund (ELBO) \cite{bib8} formulation facilitates the integration of censored data, ensuring robust estimation of time-varying risk factors and covariate dependencies. 

CR-SAVAE \cite{bib9} extends this architecture to the competing risks setting by jointly modeling multiple failure causes through a shared generative framework. Each risk is associated with its own parametric distribution, whose parameters are inferred from a common latent representation modulated by the specific event type.

\begin{figure}[ht]        
    
        \centering
        \begin{tikzpicture}[shorten >=1pt,on grid,auto] 
            \node[rectangle] (x) at (-2,0) {$x$};

            \node[draw, rectangle, fill=gray!40] (enc) [right=1 cm of x]{$\phi$};
            \node[draw, circle, fill=gray!40] (g) [right=2 cm of enc]{$z$};
            \node[rectangle] (aux) [right=2.5 cm of g]{\vdots};
            \node[draw, rectangle, fill=gray!40] (dec_cov) [above=1.5 cm of aux]{$\theta$};
            \node[draw, rectangle, fill=gray!40] (dec_r1) [above=0.5 cm of aux]{$\psi_1$};
            \node[draw, rectangle] (dec_rl) [below=0.65 cm of aux]{$\psi_K$};
            \node[draw, rectangle] (clas) [below=1.5 cm of aux]{$\omega$};
            
            \node[rectangle] (xhat) [right=2.5 cm of dec_cov] {$\hat{x}$};
          
            \node[rectangle] (that1) [right=2.5 cm of dec_r1] {$\hat{t}_1$};
            \node [right=5 cm of g] {$\vdots$}; 
            \node[rectangle] (thatk) [right=2.5 cm of dec_rl]  {$\hat{t}_K$};
            \node[rectangle] (yhat) [right=2.5 cm of clas]  {$\hat{y}$};

            \draw[->] (dec_cov) -- (xhat) node[midway, above] {$p_\theta(x|z)$}; 
            \draw[->] (dec_r1) -- (that1) node[midway, above] {$p_\psi(t_1|z)$};
            \draw[->] (dec_rl) -- (thatk) node[midway, above] {$p_\psi(t_k|z)$};
            \draw[->] (clas) -- (yhat) node[midway, above] {$p_\omega(y|z)$};
            \draw[->] (enc) -- (g) node[midway, above] {$q_\phi(z|x)$};; 

             % Encoders and decoders
            %\path[->, dotted, bend left=45] (x) edge node[midway, above] {} (g); 
            \path[->] (x) edge (enc);
            \path[->] (enc) edge (g);
            \path[->] (g) edge (dec_cov);
            \path[->] (g) edge (dec_r1);
            \path[->] (g) edge (dec_rl);
            \path[->] (g) edge (clas);
            \end{tikzpicture}
        \caption{SAVAE and CR-SAVAE architecture. Circles represent latent variables, and rectangles are Neural Networks. Shaded nodes indicate components shared in both SAVAE and CR-SAVAE, while unshaded nodes represent elements specific to the competing risks extension.  The parameters $\phi, \theta, \psi$, and $\omega$ denote the weights of the neural networks used. We use a hat (\(\hat{\cdot}\)) to denote variables predicted by the model.
}
        \label{fig:cr_savae}
    
\end{figure}
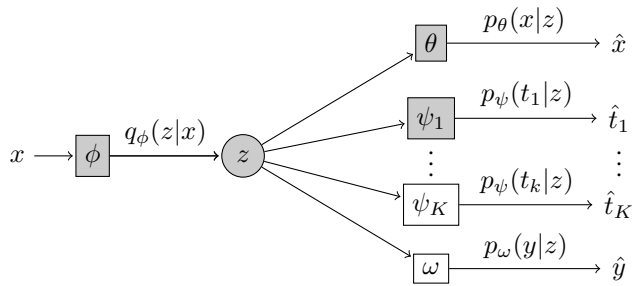

To clarify the probabilistic structure underlying these models, Figure \ref{fig:cr_savae} presents the neural network architecture used in both SAVAE and CR-SAVAE, adapted from \cite{bib9}. To facilitate understanding of the notation, we provide an appendix summarizing all symbols and definitions used throughout the paper (see Appendix \ref{secA2}). The encoder and decoder modules, implemented as multilayer perceptrons (MLPs), are tailored to handle tabular biomedical data. The architecture maps input features into a latent representation, which is then used to reconstruct the input and estimate event times.

 The encoder function, denoted by $\phi$, maps the input to the latent space. The decoder $\theta$ reconstructs the observed covariates from the latent representation. For survival modeling, $\psi$ represents the time predictor(s) that generate the parameters of the time-to-event distribution(s). In the standard survival setting, there is a single-time predictor $\psi$, as only one type of event is modeled. In contrast, in the competing risks setting, there are multiple time predictors $\psi_k$ — one for each event type — each responsible for modeling the distribution of time to a specific cause of failure. Additionally, $\omega$ denotes the event classifier, which predicts the most likely cause of failure among competing risks.
 
In SAVAE \cite{bib7}, the proposed ELBO formulation is:

\begin{equation}\label{eq:elbo}
\begin{split}
    \hat{L}(x, \theta, \psi, \phi) = 
    - \frac{1}{N} \sum_{i=1}^\Npatients \big(  & - \mathrm{D}_{\mathrm{KL}}
    (q_\phi(z|\covariablesi) \| p(z))
     + \log p_{\psi}(\timei |\reparam)   \\
     &+ \sum_{l=1}^\Ncovariates \log p_{\theta}(x^l_i | \reparam) 
    \big)
\end{split}    
\end{equation}

This formulation includes the Kullback-Leibler (KL) divergence term (denoted as $ \mathrm{D}_{\mathrm{KL}}$). The KL term acts as a regularizer, pushing the approximation of the posterior towards the prior. The latent variable \(z\) is sampled using the reparameterization trick as $z =\reparam$, which includes both the input data $x_{i}$ and the noise \(\epsilon_{i} \sim \mathcal{N}(0, I)\) is a standard Gaussian random variable. The loss also includes the time prediction term, which models the likelihood of the observed time-to-event \(t_i\), and the covariates reconstruction term, which ensures accurate reconstruction of each input feature \(x_i^l\).  In the competing risks setting, this loss formulation is extended by maintaining the same KL divergence term and covariate reconstruction loss, but adapting the time-to-event component to account for the presence of multiple possible events. Specifically, CR-SAVAE estimates a separate time distribution for each cause, and additionally includes an event prediction term to identify the most likely event type. This event prediction branch was not required in the original SAVAE, where only a single risk was considered.

Following the classical formulation in survival analysis \cite{bib21, bib64}, the contribution of each individual \(i\) to the likelihood function under censoring is expressed as:
\begin{align}
p_{\psi}(t_i | \reparam) = h(t_i | \reparam)^{\eventi} S(t_i | \reparam),
\end{align}

where $\eventi$ is the censoring indicator, defined as $\eventi = 0  \text{ if censored}$ and $\eventi = 1  \text{ if event experienced}.$ Here, $h(t_i | \reparam)$ denotes the hazard function, which quantifies the instantaneous rate of event occurrence at time $ t_i$, given that the individual has survived (i.e., has not yet experienced the event) up to that moment. 

This formulation ensures that the hazard function term is only considered when the data are not censored ($y \neq 0$), and it has also been employed in more recent frameworks such as SAVAE. In this work, the Weibull distribution is used for modelling time-to-event data, although any time distribution could be used instead, given that its log-likelihood is differentiable.

\subsection{SAMVAE Architecture for Single and Competing Risks} \label{samvae}
\begin{figure}[ht]
\centering
\begin{tikzpicture}[node distance=1.1cm and 1.5cm, on grid, auto]

% Primera fila
\node[rectangle] (x1) at (0,2.5) {$x^{\langle 1 \rangle}
$};
\node[rectangle, draw, fill=gray!40, minimum width=0.8cm, minimum height=0.5cm] (e1) [right=of x1] {{$\phi^{\langle 1 \rangle}$}};
\node[circle, draw, fill=gray!40, minimum size=0.7cm] (z1) [right=of e1] {$z^{\langle 1 \rangle}
$};

% Segunda fila
\node[rectangle] (x2) at (0,0.5) {$x^{\langle 2 \rangle}
$};
\node[rectangle, draw, fill=blue!10, minimum width=0.8cm, minimum height=0.5cm] (e2) [right=of x2] {{$\phi^{\langle 2 \rangle}$}};
\node[circle, draw, fill=blue!10, minimum size=0.7cm] (z2) [right=of e2] {$z^{\langle 2 \rangle}$};

% Puntos suspensivos
\node at (0,-0.4) {$\vdots$};
\node at (1.5,-0.4) {$\vdots$};
\node at (3,-0.4) {$\vdots$};

% Tercera fila
\node[rectangle] (xM) at (0,-1.5) {$x^{\langle M \rangle}
$};
\node[rectangle, draw, fill=blue!10, minimum width=0.8cm, minimum height=0.5cm] (eM) [right=of xM] {{$\phi^{\langle M \rangle}$}};
\node[circle, draw,  fill=blue!10, minimum size=0.7cm] (zM) [right=of eM] {$z^{\langle M \rangle}$};

\draw[->] (x1) -- (e1);
\draw[->] (e1) -- (z1);
\draw[->] (x2) -- (e2);
\draw[->] (e2) -- (z2);
\draw[->] (xM) -- (eM);
\draw[->] (eM) -- (zM);

% Caja de concatenación
\node[rectangle, draw, thick, minimum width=0.8cm, minimum height=5.5cm, fill=gray!40] (concat) at (5,0.5) {z};

% Flechas z -> concat
\draw[->] (z1) -- (concat);
\draw[->] (z2) -- (concat);
\draw[->] (zM) -- (concat);

% Decodificadores para D_1, ..., D_M
\node[rectangle, draw, minimum width=1.2cm, minimum height=0.5cm, fill=gray!40] (d1) at (8,3.2) {$\theta^{\langle 1 \rangle}$}; 

\node[rectangle, draw, minimum width=1.2cm, minimum height=0.5cm, fill=blue!10] (d2) at (8,2.4) {$\theta^{\langle 2 \rangle}$}; 

\node at (8,1.9) {$\vdots$}; 

\node[rectangle, draw, minimum width=1.2cm, minimum height=0.5cm, fill=blue!10] (dm) at (8,1.2) {$\theta^{\langle M \rangle}$}; 

% Clasificador
\node[rectangle, draw, minimum width=1.2cm, minimum height=0.5cm, fill=gray!40] (t1) at (8,0.3) {$\psi_{ 1 }$};

\node at (8,-0.2) {$\vdots$}; 

\node[rectangle, draw, minimum width=1.2cm, minimum height=0.5cm] (tK) at (8,-0.9) {$\psi_{K }$};
\node[rectangle, draw, minimum width=1.2cm, minimum height=0.5cm] (cls) at (8,-1.8) {$\omega$};

% Flechas concat -> decodificadores
\draw[->] (concat) -- (d1); 
\draw[->] (concat) -- (d2); 
\draw[->] (concat) -- (dm); 
\draw[->] (concat) -- (t1);
\draw[->] (concat) -- (tK);
\draw[->] (concat) -- (cls);

\node[rectangle] (xhat1) at (10,3.2) {$\hat{x}^{\langle 1 \rangle}$};
\node[rectangle] (xhat2) at (10,2.4) {$\hat{x}^{\langle 2 \rangle}$};
\node at (10,1.9) {$\vdots$}; 
\node[rectangle] (xhatm) at (10,1.2) {$\hat{x}^{\langle M \rangle}$};
\node[rectangle] (that1) at (10,0.3) {$\hat{t}_{1}$};
\node at (10,-0.2) {$\vdots$}; 
\node[rectangle] (thatk) at (10,-0.9) {$\hat{t}_{K}$};
\node[rectangle] (yhat) at (10,-1.8) {$\hat{y}$};

\draw[->] (d1) -- (xhat1);
\draw[->] (d2) -- (xhat2);
\draw[->] (dm) -- (xhatm);
\draw[->] (t1) -- (that1);
\draw[->] (tK) -- (thatk);
\draw[->] (cls) -- (yhat);

\end{tikzpicture}
\caption{ General architecture of SAMVAE. Gray-shaded nodes denote components shared with both SAVAE and CR-SAVAE. Unshaded nodes are specific to the CR-SAVAE, while the remaining colored nodes represent novel additions introduced by SAMVAE. Each input modality \(x^{\langle m \rangle}
\) is encoded via its own encoder of parameters $\phi^{\langle m \rangle}$ into a latent representation \(z^{\langle m \rangle}
\). The concatenated latent vector is used to decode covariates \(\theta\), survival time(s) \(\psi\), and (if competing risks) event type \(\omega\).}
\label{fig:samvaearq}
\end{figure}
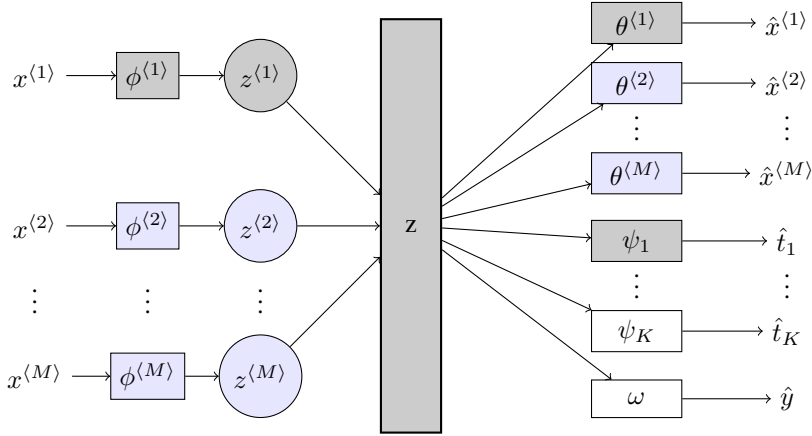

We now introduce SAMVAE, our proposed multimodal extension of the two previously described architectures, designed to address both standard survival analysis (when a single event of interest is present in the dataset) and competing risks scenarios (which involve the presence of multiple mutually exclusive event types). SAMVAE is capable of integrating heterogeneous data modalities, including both tabular data (e.g., clinical and omic modalities) and image data. It leverages a shared latent representation learned from all available modalities, enabling joint modeling of heterogeneous inputs for time-to-event prediction. Figure \ref{fig:samvaearq} illustrates the architecture adapted for our proposed SAMVAE framework. Each input modality $x^{\langle m \rangle}
$ is first encoded into a modality-specific latent representation $z^{\langle m \rangle}$ via a dedicated encoder $\phi^{\langle m \rangle}$. These latent vectors are subsequently concatenated into a unified latent representation $z$.

 The shaded blue components highlight the novel contributions introduced in SAMVAE, including decoders and predictive branches specifically tailored for temporal and clinical outcomes. Meanwhile, the remaining gray-shaded components correspond to SAVAE and CR-SAVAE preserved from the base architecture.

The proposed model involves as many Deep Neural Networks (DNNs) as the number of encoders and decoders required, depending on the number of modalities involved. An additional module is trained for time-to-event prediction. These decoders are designed to infer covariates, while a dedicated predictor is trained to estimate the time parameters, respectively. Specifically, for $M$ modalities, the total number of DNNs in the model depends on the type of survival task. In a survival analysis scenario, the model uses $2M + 1$ networks: $M$ encoders, $M$ decoders for reconstructing covariates, and one dedicated to time-to-event prediction. In the case of competing risks, the model requires $2M + K + 1$ networks. This includes $M$ encoders, $M$ covariate decoders, $K$ separate predictors for time-to-event parameters (one predictor per risk), and one additional network acting as a classifier to predict which event (risk) will occur.

In deep learning, architectural choices are guided by data characteristics and the inductive biases of neural networks \cite{bib12}. Given the multimodal nature of the dataset, comprising tabular (clinical and omic) and image data, distinct encoder–decoder architectures are employed for each modality: MLPs for tabular data and Convolutional Neural Networks (CNNs) for image data.

The latent spaces $z^{\langle m \rangle}$ generated by each encoder are concatenated into a unified latent representation $z$, resulting in an approximate posterior  $q_\phi(z|x) = q_\phi^{\langle 1 \rangle}(z^{\langle 1 \rangle}|x^{\langle 1 \rangle}) \, q_\phi^{\langle 2 \rangle}(z^{\langle 2 \rangle}|x^{\langle 2 \rangle}) \cdots q_\phi^{\langle M \rangle}(z^{\langle M \rangle}|x^{\langle M \rangle})$, where the joint posterior is defined by the concatenation of all modality-specific latent distributions.
 This shared latent space $z$ serves as the foundation for generating samples that are fed into all decoders. By ensuring that each decoder receives information from all modalities, the model effectively integrates complementary data sources. This is particularly crucial for the time-to-event predictor(s), as it enables survival analysis and competing risks predictions to be informed by the most relevant modalities, since it estimates the time-to-event distribution parameters from the shared latent representation.

In the present work, the Weibull distribution is adopted for modeling time-to-event data. However, any time distribution with a differentiable log-likelihood could be used instead, including those defined over discrete time. Although the proposed approach is compatible with both continuous and discrete time settings, we restrict our experiments to the continuous Weibull distribution due to its widespread use in survival analysis \cite{bib70}, its demonstrated effectiveness in prior studies \cite{bib57}, and the fact that it does not assume proportional hazards.

Building upon the formulation presented in the Equation \ref{eq:elbo}, where the ELBO explicitly accounts for each dimension of the covariates vector, the proposed extension adapts this framework to handle multimodal data. Specifically, since the model includes as many encoders/decoders as there are modalities, the ELBO is modified to include an additional summation over the number of modalities. This additional summation ensures that each modality contributes appropriately to the overall objective, resulting in the following ELBO formulation (Equation \ref{elbosamvae}). Additionally, SAMVAE operates in a parametric manner: both the time and the covariates can follow any distribution that has a differentiable log-likelihood function, which highlights its flexibility.

\begin{align}
    \hat{L}(x, \theta, \psi, \phi) = - \frac{1}{N} \sum_{i=1}^N \bigg( 
    &- \mathrm{D}_{\mathrm{KL}}(q_\phi(z|\covariablesi) \| p(z)) 
    + \log p_{\psi}(\timei |\reparam) \notag  \nonumber\\
    &+  \sum_{m=1}^M \sum_{l_m=1}^{\Ncovariates_m} \log p_{\theta}(\covariablesi^{l_m} | g_{\phi}(x_{i}, \epsilon_i)) 
    \bigg)
\label{elbosamvae}    
\end{align}

In this formulation:  
\begin{itemize}
    \item \( \mathrm{D}_{\mathrm{KL}}(q_\phi(z|x_i) \| p(z))\): This term represents the KL divergence between the approximate posterior \(q_\phi(z|x_i)\) and the prior distribution \(p(z)\), where $z$ denotes the fused latent representation obtained from the concatenation of all modality-specific latent vectors $z^{\langle m \rangle}$. As a regularization term, it encourages the aggregated latent variable $z$ to remain close to the prior distribution.
    \item \(\log p_{\psi}(t_i | g_\phi(x_i, \epsilon_i))\): This term, also known as the time prediction term, corresponds to the likelihood of the time-to-event variable \(t_i\), given the deterministic transformation \(g_\phi(x_i, \epsilon_i)\), which includes both the input data \(x_i\) and the noise \(\epsilon_i\). The function $g_\phi$ is defined as the concatenation of all the $g_{\phi}^{\langle M \rangle}$, one for each modality \(M\). 
    \item \(\sum_{m=1}^M \sum_{l_m=1}^{\Ncovariates_m} \log p_{\theta}(\covariablesi^{l_m} | g_{\phi}(x_{i}, \epsilon_i))\): This summation accounts for the reconstruction term of the covariates vector \(x_i^{l_m}\), ensuring that each dimension \(l_m\) is explicitly modelled. Here, \(M\) represents the number of modalities, and \(\Ncovariates_m\) denotes the dimensions of the covariates specific to each modality \(m\).
\end{itemize}

Similarly, the total loss used for training SAMVAE is adapted to account for multiple event types. In this setting, the censoring indicator is defined as \eventi = 0 for censored patients, and \eventi $\neq$ 0 for uncensored patients who experienced event type $k$. This generalization allows the model to distinguish between competing outcomes during training and adjust the loss function accordingly. Therefore, in SAMVAE we define the loss as in:

\begin{align}
\hat{L}(x, \theta, \psi, \phi, \omega) = -\frac{1}{\Npatients} \sum_{i=1}^{\Npatients} \bigg[ 
    & \, -{D}_{\text{KL}}(q_\phi(z \mid x_i) \,\|\, p(z)) \nonumber 
     + \sum_{m=1}^M \sum_{l_{m} = 1}^{L_m} \log p_{\theta}(x_i^{l_m} \mid  g_{\phi}(x_{i}, \epsilon_i)) \nonumber \\
    &    + \mathbb{I}(\eventi \neq 0) \bigg( \log p_{\psi}(t_i \mid  \reparam) + \log p_\omega(\eventi \mid \reparam)  \nonumber \bigg)\\
    & + \mathbb{I}(\eventi = 0) \log\left(1 - \sum_{k=1}^{\Nevents} \text{CIF}_k(t_i \mid \reparam)\right)   
\bigg]
\label{eq:CRS-loss}
\end{align}

This formulation extends the original CR-SAVAE loss by incorporating terms specific to time-to-event modeling and competing risks, while accounting for the joint contribution of multiple data modalities in the latent representation. Once the conditional survival functions $P(T \leq t \mid Y=k, x)$ and risk probabilities $P(Y=k \mid x)$ are estimated, the CIFs can be computed as in Equation \ref{eq:CIF}.

To ensure that the latent representation $z$ is meaningful, we include a reconstruction term in the loss function. While it would be possible to rely solely on the encoder to obtain $z$ for downstream survival prediction, the inclusion of a decoder and the associated reconstruction loss imposes an additional constraint that enforces $z$ to serve as a compact and informative summary of each patient's multimodal data. 

 In this formulation:
 \begin{itemize}
    \item \textbf{KL Divergence Term:} ${D}_{\text{KL}}(q_\phi(z\mid x_i) \,\|\, p(z))$ penalizes deviations between the approximate posterior and the prior over the latent variables, encouraging a regularized and informative latent space.
    
    \item \textbf{Reconstruction Term:} $\sum_{m=1}^M \sum_{l_{m} = 1}^{L_m} \log p_{\theta}(x_i^{l_m} \mid g_{\phi}(x_{i}, \epsilon_i))$ captures the multimodal reconstruction loss, where each modality $m$ and its $L_m$ constituent features are reconstructed from the shared latent representation $z_i$. This term ensures that the model learns meaningful representations of all input modalities. 

    \item \textbf{Event-Specific Likelihood (Uncensored Cases):} For patients who experienced an event ($\eventi \neq 0$), the loss includes two terms: $\log p_{\psi}(t_i \mid \reparam)$ models the time-to-event given that the event type information provided by the latent representation, and $\log p_\omega(\eventi \mid \reparam)$ estimates the probability of observing the specific event type. These terms guide the model to predict both when and what kind of event is likely to occur. 

    \item \textbf{Censoring Term (Censored Cases):} For censored patients ($\eventi = 0$), the loss includes the logarithm of the survival probability across all event types: $\log\left(1 - \sum_{k=1}^{\Nevents} \text{CIF}_k(t_i \mid \reparam)\right)$. This term ensures that the model correctly accounts for censoring by maximizing the likelihood of survival beyond the observed time.

\end{itemize}

For the reconstruction of tabular data, the same log-likelihood functions as those used in SAVAE are employed, corresponding to Gaussian, Bernoulli, and Categorical distributions depending on the variable type. In the case of image data we use the Mean Squared Error (MSE), which corresponds to a Gaussian likelihood with fixed variance. The total reconstruction loss is obtained by summing the losses across modalities. This additive formulation could be further improved by introducing modality-specific weights, allowing the model to balance their relative contributions more effectively, as suggested in \cite{bib11}.

\section{Experimental Analysis}\label{sec4}

This section introduces the experimental validation framework designed to assess the performance of the proposed SAMVAE model. The goal is to determine its ability to integrate heterogeneous medical data for survival prediction—both in standard survival analysis and competing risks—and to compare it with baseline and state-of-the-art methods under fair conditions.

We work with multimodal patient data that includes tabular information (clinical variables and omics data) and histopathological images in the form of patch-level representations. Each modality differs in dimensionality and data structure. 

The ultimate goal of this study is to compare the proposed SAMVAE model with baseline and state-of-the-art approaches, including configurations based on clinical, omics, and imaging modalities. Before performing such comparisons, we first identify the optimal configuration of SAMVAE for each cancer type and data modality. Since the encoders produce a Gaussian latent space, the dimensionality of this latent representation and the number of neurons in each hidden layer constitute the key hyperparameters to be tuned. We perform a grid search over these hyperparameters by exploring multiple configurations, aiming to identify the setup that maximizes model performance.

After selecting the optimal parameters for each modality, we evaluate different multimodal combinations—including various omics modalities and varying numbers of histopathological patches per patient—to determine the best integration strategy. The final selected configuration, integrating clinical, omics, and imaging data, is then used in the main experiments and comparative evaluation against baseline and state-of-the-art methods.

\subsection{Datasets and Preprocessing}

TCGA provides a comprehensive repository of genomic and clinical data from over 11,000 patients across 33 cancer types \cite{bib10}. In this study, we selected two representative cohorts, Breast Invasive Carcinoma (BRCA) and Lower Grade Glioma (LGG), due to the availability of multimodal and publicly accessible patient data. For each cohort, we used two datasets: one for standard survival analysis with a single risk (BRCA-SA and LGG-SA) and another for competing risks modeling (BRCA-CR and LGG-CR).

For the competing risks analysis, we utilized the time-to-event and event status annotations provided for the Progression-Free Interval (PFI) endpoint, as recommended by TCGA guidelines \cite{bib24}. PFI captures the duration from the date of diagnosis until the first occurrence of a new tumor event, which includes progression, recurrence, or death with evidence of disease. Specifically, it distinguishes between three types of outcomes: 1 for the event of interest (tumor progression or cancer-related death), 2 for competing risks (death without prior progression), and 0 for censored observations. In this study, we applied this competing risks framework to the BRCA-SA and LGG-SA cohorts, utilizing their respective competing risks datasets (BRCA-CR and LGG-CR).

A total of six data modalities were used: (i) clinical variables (tabular), (ii)DNAm, (iii) CNV, (iv) miRNA, (v) RNAseq, and (vi) WSIs. These modalities span three major categories of healthcare data—clinical, molecular, and imaging—commonly employed in multimodal biomedical studies \cite{bib5}.

Tailored preprocessing strategies were applied to each data modality, taking into account its specific characteristics and underlying data structure. For omics data, due to their high dimensionality, reduction techniques such as the minimal-redundancy-maximal-relevance (mRMR) criterion \cite{bib11}, Principal Component Analysis (PCA), retaining the top 100 components, and filtering by the most frequently mutated genes were applied, following practices established in works like ConcatAE/CrossAE \cite{bib16} and Multisurv \cite{bib18}. For WSIs, we used the Clustering-constrained Attention Multiple Instance Learning (CLAM) framework \cite{bib17}, which provides predefined preprocessing configurations for TCGA data for the tissue segmentation and extraction of the most informative patches. Each WSI is divided into smaller patches, which are subsequently ranked by relevance using CLAM's attention-based mechanism. For each patient, we select the top informative patches and use them as input to our model.

Given the inherent heterogeneity of these modalities, each modality required specific preprocessing steps (fully documented in the GitHub repository \cite{bib11}). However, our primary focus lies in designing a flexible and scalable multimodal integration strategy rather than preprocessing itself. Although the current preprocessing pipeline is effective, it remains adaptable to potential future improvements.
\subsection{Performance metrics}
 
In survival analysis, the Concordance Index (C-index) is a commonly used metric to measure the quality of a predictive model in ranking individuals by their risk of experiencing an event (e.g., death, disease recurrence). Formally, for a given time \( t \) and two patients \( i \), \( j \), the time-dependent C-index can be expressed as \cite{bib65}:

\begin{equation}
C_{\text{\textit{index}}}(t) \approx 
\frac{\sum_{i \ne j} A_{i,j} \cdot \mathbb{I}\left( F(t \mid x_i) > F(t \mid x_j) \right)}{\sum_{i \ne j} A_{i,j}(t)} 
\end{equation}

where \( F(t \mid x) \) is the estimated cumulative distribution function for time-to-event and $A_{i,j}(t) = \mathbb{I}\left(Y_i = 1|T_i < T_j,\; T_i \le t\right)$. A C-index close to 1 indicates better alignment between predicted risk and observed outcomes, while a value of 0.5 corresponds to random predictions.

In competing risks settings, the C-index for event $k$ measures the concordance between the predicted CIF for a patient at the time of the event and that of patients who have not yet experienced the event, considering event timing and censoring. Following the definition in \cite{bib9}, it is computed as:

\begin{equation}
C_{\text{\textit{index}}} (t) \approx 
\frac{
\sum_{i \ne j} A_{k,i,j} \cdot \mathbb{I} \left( \mathrm{CIF}_k(t \mid x_i)
 \leq \text{CIF}_k(t | x_j) \right)
}{
\sum_{i \ne j} A_{k,i,j}
},
\label{eq:cindex}
\end{equation}

where $A_{k,i,j} = \mathbb{I}\left(Y_i = K| T_i < T_j,\; T_i \le t\right)$ is an indicator function for whether the pair of patients $(i, j)$ is comparable for event $k$, and $\mathbb{I}(\cdot)$ is the indicator function.

The IBS quantifies the squared error between predicted survival probabilities and actual outcomes, accounting for censoring via Inverse Probability of Censoring Weighting (IPCW) \cite{bib14}. The Brier Score (BS) at time $t$ is defined as in \cite{bib7}:

\begin{align}
BS(t) = \frac{1}{N} \sum_{i=1}^N \Bigg[ \frac{\big(S(t \mid x_{i})\big)^2}{G(t_{i})} \cdot \mathbb{I}(t_{i} < t, y_{i} = 1) + \frac{\big(1 - S(t \mid x_{i})\big)^2}{G(t)} \cdot \mathbb{I}(t_{i} \geq t) \Bigg],
\end{align}

where $S(t \mid x_i)$ is the predicted survival probability,  $\mathbb{I}(\cdot)$ is an indicator function, and $G(t)$ is the censoring survival function.  $G(t)$ represents the estimated probability of not being censored at time \(t\), and is obtained using the Kaplan-Meier (KM) estimator \cite{bib21}.

The BS reflects both discrimination and calibration. To summarize performance over time, the Integrated Brier Score (IBS) averages the BS across the follow-up period as:

\begin{align}
IBS(t_{\textit{max}}) = \frac{1}{t_{\textit{max}}} \int_0^{t_{\textit{max}}} BS(t) dt,
\end{align}

with $t_{\text{max}}$ denoting the maximum follow-up time.
 
As discussed in \cite{bib9}, in the context of competing risks, the IBS quantifies the overall discrepancy between the predicted and observed CIFs over time, while accounting for the presence of multiple events. The time-dependent Brier Score \( BS(t) \) is given by:

\begin{align}
BS_{k}(t) = \frac{1}{N} \sum_{i=1}^N \left[ \frac{(1 - CIF_k(t \mid x_{i}))^2}{G(t_{i})} \cdot \mathbb{I}(t_{i} < t, y_{i} \neq 0) + \frac{(CIF_k(t \mid x_{i}))^2}{G(t)} \cdot \mathbb{I}(t_{i} \geq t) \right].
\end{align}

 \subsection{Parameter and Modality Selection}

    For each modality, the optimal parameter combination was identified using five random seeds to ensure the robustness and stability of the selection process. The optimized parameters were the latent dimension and the hidden size (values tested for each modality and configuration are detailed in the Appendix \ref{secA1}, see Table \ref{values_tested}). Once the best set of parameters was found, we evaluated the combination of omics and clinical data that yielded the best performance. Similarly, results were analyzed using the combination of clinical data with a different number (1, 5, 10, or 15) of the most informative patches from WSIs per patient, extracted using the CLAM model. 
    
    For the final stage of our experimental pipeline—after identifying the best omics configuration, optimal patch selection, and incorporating clinical data—we conducted training using 10 different random seeds to ensure the robustness and reliability of the results. These are the experiments presented in this paper as the main evaluation of the proposed SAMVAE model.

 Given the sensitivity of VAE architectures to initialization conditions,  our model includes specific design choices to ensure robust and representative evaluation. The C-index and IBS are averaged over the three best-performing seeds. Both values are then used to calculate \( p \)-values by performing a unilateral t-test, ensuring statistical significance in the results.
 
The final selection of the best hyperparameter combination is based on a two-stage criterion:

\begin{itemize}
    \item We applied the Holm-Bonferroni (HB) \cite{bib19} correction to control for Type I errors due to multiple testing. This method is more powerful than the traditional Bonferroni correction and well-suited to our analysis as it does not assume independence between tests. Combinations where the corrected \( p \)-values of the C-index and IBS do not exceed the threshold of 0.01 are discarded.

\item Among the remaining combinations, the one that maximizes the CI-IBS (C-index minus Integrated Brier Score) metric is chosen. This metric is selected because a higher C-index indicates better performance, while a lower IBS is preferable; therefore, the IBS is subtracted. By combining them into a single metric, we obtain a balanced assessment of the model's performance.
\item In the case of competing risks datasets, the combined metric is computed as the average of the C-index values across all considered risks, minus the average of their respective IBS values. This metric is used to identify the hyperparameter combination that optimizes overall performance across risks.

\end{itemize}

Therefore, once the best hyperparameters have been identified for each dataset and scenario, and the optimal combination of omics data and image patch number has been determined, these are combined with the clinical modality to establish the final configuration used for the evaluation of SAMVAE and its comparison with state-of-the-art methods.

 \section{Results} 
 The experiments reported in this section correspond to the configurations that combine the inclusion of clinical data, the best-performing omics modality (or modalities), and the optimal number of image patches. These configurations represent the most effective fusion strategies identified through hyperparameter tuning and intermediate experiments conducted for each dataset. These final multimodal combinations form the basis for evaluating the performance of SAMVAE.

\subsection{Multimodal Survival Analysis}\label{resultssa}
This section presents the results obtained from the survival analysis tasks in the LGG-SA and BRCA-SA datasets. In addition, the performance of SAMVAE on BRCA-SA is compared with state-of-the-art approaches presented in Section \ref{sec2_31} for multimodal integration of clinical, omics, and imaging data. Table \ref{tabla:lgg} presents the results of the final multi-modal experiments for the LGG-SA cohort. It compares different combinations of clinical, DNAm, and histological image data (using 10 patches) in terms of survival prediction performance. The selection of DNAm, as the omic modality, and the use of 10 patches for histological data is based on intermediate experiments (detailed in the previous section and exemplified by the tables shown in Appendix \ref{secA3}).

\renewcommand{\arraystretch}{1.5}  
\begin{table}[ht]
\centering

% Subtabla LGG
\begin{subtable}{\textwidth}
\centering
\resizebox{0.9\textwidth}{!}{
\begin{tabular}{|c|c|c|c|c|c|c|c|}

\hline
\textbf{Clinical} & \textbf{DNA Methylation} & \textbf{10 Patches}  & \textbf{C-index} & \textbf{IBS} & \textbf{CI - IBS} & \textbf{p-CI HB} & \textbf{p-IBS HB} \\
\hline
\textbullet &   &  &\textbf{ 0.773 ± 0.089 }& \textbf{0.201 ± 0.035} & \textbf{0.572 ± 0.084} &       \textbf{  1.000}  &         \textbf{0.591} \\
\hline
  & \textbullet &       & 0.615 ± 0.042 & 0.196 ± 0.014 & 0.420 ± 0.049 &           $<\!1e^{-3}$ &         0.639 \\
\hline
  &   & \textbullet & 0.592 ± 0.029 & 0.195 ± 0.015 & 0.397 ± 0.037 &          $<\!1e^{-3}$    &         0.773 \\
\hline
\textbullet & \textbullet &        & \textbf{0.754 ± 0.050} & \textbf{0.190 ± 0.027} & \textbf{0.565 ± 0.060} &         \textbf{1.000} &        \textbf{ 1.000  }   \\  
\hline
\textbullet &   & \textbullet      & \textbf{0.773 ± 0.043} &\textbf{ 0.187 ± 0.020 }&\textbf{ 0.586 ± 0.050} &        \textbf{  1.000} &       \textbf{  1.000 }    \\ 
\hline
  & \textbullet & \textbullet &   0.620 ± 0.059 & 0.198 ± 0.019 & 0.421 ± 0.071 &        $<\!1e^{-3}$     &         0.447 \\
\hline
  \textbullet & \textbullet &   \textbullet   & \textbf{0.761 ± 0.047} & \textbf{0.186 ± 0.023} &\textbf{ 0.576 ± 0.052} &       \textbf{  1.000}&        \textbf{ 1.000}    \\
\hline
\end{tabular}
}
\subcaption{Performance metrics for different modality combinations on LGG-SA.}
\label{tabla:lgg}
\end{subtable}

\vspace{1em} % espacio vertical entre tablas

% Subtabla BRCA
\begin{subtable}{\textwidth}
\centering
\resizebox{0.9\textwidth}{!}{
\begin{tabular}{|c|c|c|c|c|c|c|c|}

\hline
\textbf{Clinical} & \textbf{CNV} & \textbf{15 Patches}  & \textbf{C-index} & \textbf{IBS} & \textbf{CI - IBS}  & \textbf{p-CI HB} & \textbf{p-IBS HB} \\
\hline
\textbullet &   &  & \textbf{0.722 ± 0.057 }& \textbf{0.175 ± 0.025} & \textbf{0.546 ± 0.066 } &        \textbf{1.000}   &         \textbf{1.000 }  \\
\hline
  & \textbullet &        & 0.636 ± 0.034 & 0.195 ± 0.020 & 0.441 ± 0.031 &          $<\!1e^{-3}$    &         0.099 \\
\hline
  &   & \textbullet  & 0.615 ± 0.074 & 0.203 ± 0.029 & 0.411 ± 0.085 & 0.001 &         0.034 \\
\hline
\textbullet & \textbullet &         & \textbf{0.705 ± 0.061} & \textbf{0.191 ± 0.035} & \textbf{0.514 ± 0.077} &       \textbf{1.000} &        \textbf{ 0.602} \\
\hline
\textbullet &   & \textbullet     & \textbf{0.699 ± 0.046} & \textbf{0.192 ± 0.019} & \textbf{0.507 ± 0.054} &         \textbf{0.922} &       \textbf{  0.195} \\
\hline
  & \textbullet & \textbullet  & 0.640 ± 0.060 & 0.196 ± 0.021 & 0.444 ± 0.059 &      0.003&         0.094 \\ 
\hline
  \textbullet & \textbullet &   \textbullet    & \textbf{0.688 ± 0.060} & \textbf{0.192 ± 0.019} & \textbf{0.496 ± 0.057 }&          \textbf{ 0.504 }&         \textbf{0.178} \\
\hline
\end{tabular}
}
\subcaption{Performance metrics for different modality combinations on BRCA-SA.}
\label{tabla:brca}
\end{subtable}

\caption{Both tables report the C-index, IBS, and their difference (CI-IBS). P-values, corrected using HB, are computed by comparing each combination against the best-performing one in terms of C-index and IBS, respectively. Bold values indicate combinations whose performance is not significantly different (p-values $>$0.01) from the best one, for both C-index and IBS.}
\label{fig:combined_tables}
\end{table}  

While clinical data alone achieves strong predictive performance with a C-index of 0.773 ± 0.089, the combination with 10 patches preserves this discriminative power (C-index = 0.773 ± 0.043) and simultaneously improves calibration, as indicated by a lower IBS. Moreover, adding both omics and image modalities results in a slightly lower IBS (0.186 ± 0.023), showing consistent gains in model calibration through multimodal integration.  Notably, the combinations using all three modalities, clinical + patches, clinical + DNAm, and clinical data alone, exhibit similar overall performance, suggesting that multiple configurations can achieve comparable predictive accuracy and calibration.

Similarly to the experiments conducted for the LGG-SA cohort, the final multimodal configurations for the BRCA-SA dataset are presented in Table \ref{tabla:brca}. The table evaluates different combinations of clinical, CNV, and histological image data (using 15 patches) in terms of survival prediction performance (choice supported by intermediate experiments, see Appendix \ref{secA3}). As with LGG-SA, clinical variables of BRCA-SA alone achieve strong predictive performance (C-index = 0.722 ± 0.057). It is crucial to highlight that both unimodal clinical models and their combinations with other modalities consistently yield strong predictive performance. This observation underlines the value of clinical variables as a robust baseline in survival analysis. 

Established clinical evidence identifies traditional prognostic factors — particularly age — as key determinants of survival outcomes \cite{bib23}, a pattern that is reaffirmed in our results, where clinical variables clearly dominate survival prediction. Therefore, the proposed multimodal integration can be considered successful, as it not only reaches the strong discriminative performance of clinical data alone, but also improves calibration metrics in some combinations.

\begin{figure}[ht]
\centering

\begin{subfigure}[b]{0.9\textwidth}
    \centering
    \includegraphics[width=\textwidth]{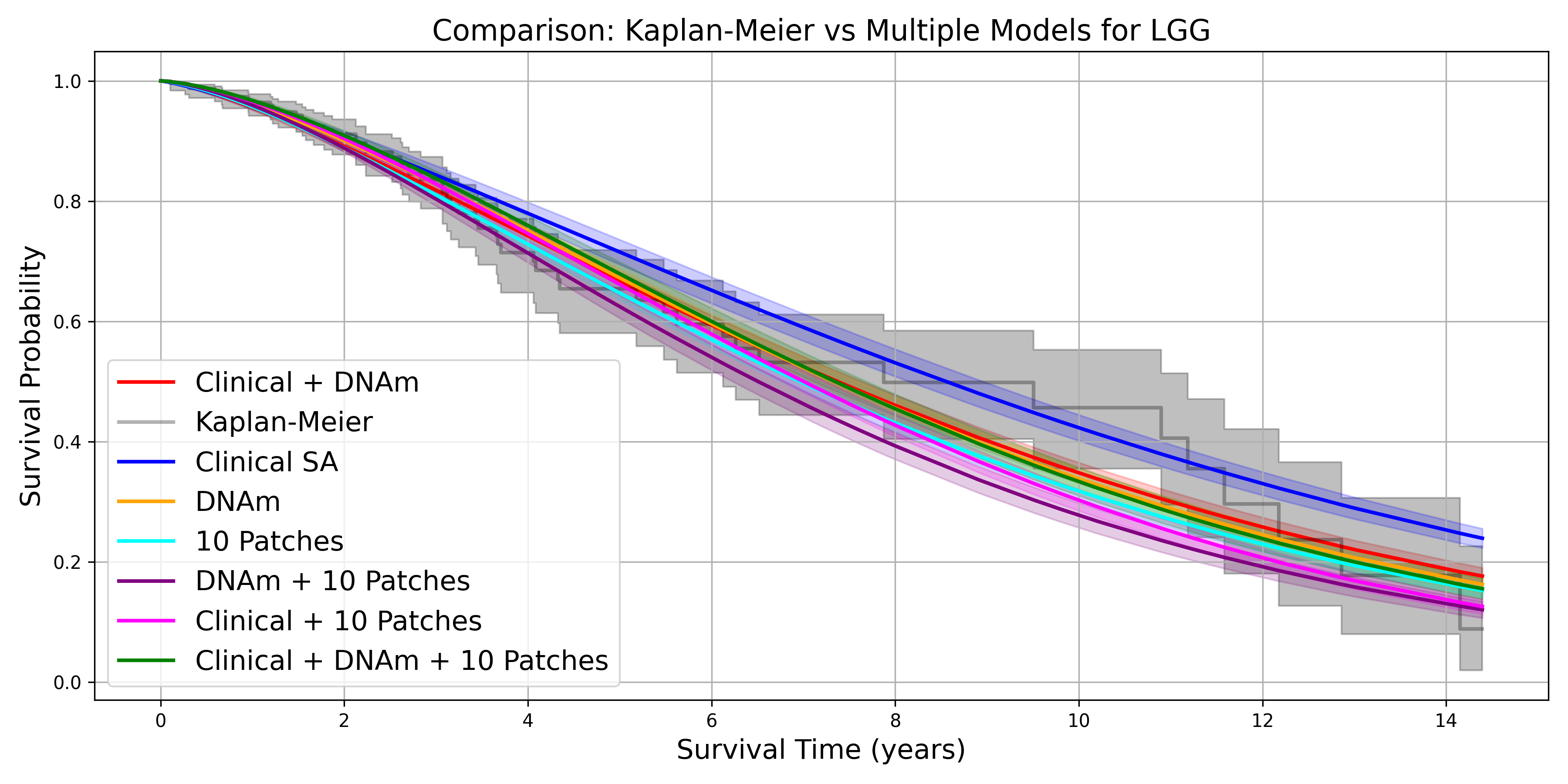}
    \subcaption{Predicted survival curves on the LGG-SA dataset.}
    \label{figLGG}
\end{subfigure}

\begin{subfigure}[b]{0.9\textwidth}
    \centering
    \includegraphics[width=\textwidth]{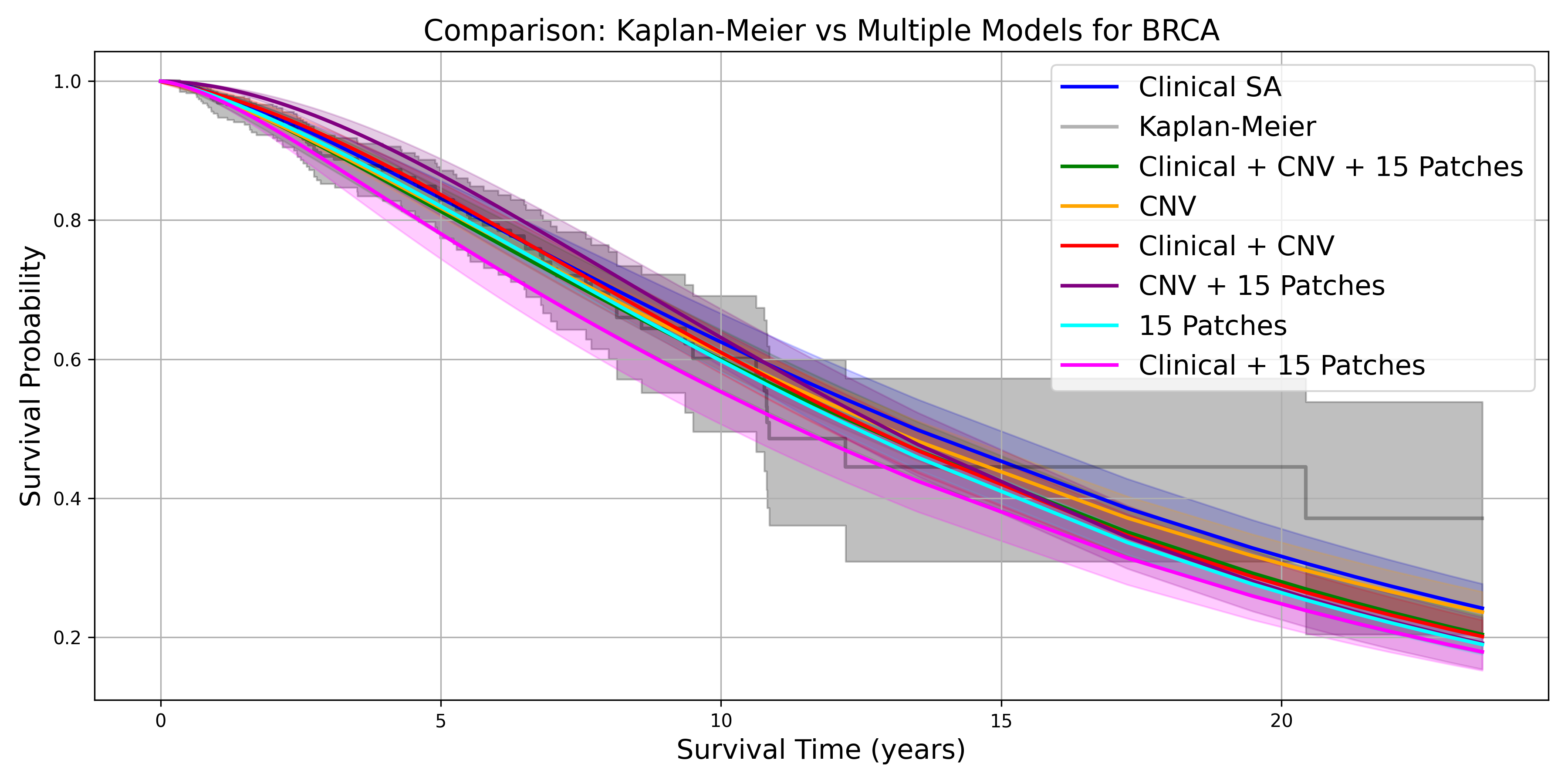}
    \subcaption{Predicted survival curves on the BRCA-SA dataset.}
    \label{fig2}
\end{subfigure}

\caption{Average predicted survival curves derived from different combinations of input modalities. Each colored line corresponds to a specific modality configuration, as indicated in the legend. The shaded gray region represents the KM estimate. The consistent overlap observed across all modality combinations
highlights the model’s effectiveness to generate accurate survival predictions.}
\label{fig:combined_survival}
\end{figure} 

To illustrate the performance of our SAMVAE model in the context of survival analysis in LGG-SA and BRCA-SA cohorts, Figures \ref{figLGG} and \ref{fig2} show the average predicted survival curves obtained from different combinations of data modalities, as indicated in the legend. The shaded gray area corresponds to the KM estimate \cite{bib21}, which serves as a non-parametric reference survival curve calculated from the observed event data. For each modality combination, the survival curve is computed as the mean of the predicted survival probabilities across all test patients. The shaded region around each curve represents the 95\% confidence interval.

As shown in Figures \ref{figLGG} and \ref{fig2}, all modality combinations — including those that exclude clinical data — yield survival curves that closely follow the KM estimates for both LGG-SA and BRCA-SA cohorts. This consistent alignment across diverse input configurations (unimodal and multimodal combinations) underscores the model’s robustness and its ability to generate accurate and well-calibrated survival predictions, regardless of the modality combination. The reliable overlap between predicted and KM survival curves demonstrates that the proposed framework can effectively and successfully capture the fundamental structure of survival events from heterogeneous and multiple modalities.

The project repository includes an interactive HTML visualization that compares the survival curves predicted by different trained models against the Kaplan-Meier estimate for LGG and BRCA, allowing users to visually assess model performance across patient groups (see Appendix \ref{sece1}, available at \href{https://albagarridolopezz.github.io/SAMVAE/interactive_plot_lgg_sa.html}{Interactive LGG Survival Plot} and \href{https://albagarridolopezz.github.io/SAMVAE/interactive_plot_brca_sa.html}{Interactive BRCA Survival Plot}).
 
 \subsubsection{Comparison with State-of-the-Art Multimodal Models}
 
Some state-of-the-art approaches, such as \cite{bib18}, report only multimodal results that include clinical data, omitting scenarios in which only omic or imaging data are available, or in which clinical information is absent. This lack of unimodal and ablation baselines limits the interpretability of the reported performance and complicates a precise understanding of the individual contribution of each modality. Our work addresses this gap by providing a transparent and systematic evaluation across all modality combinations, demonstrating that clinical data alone already achieves strong performance, a necessary reference point to fairly assess the added value of multimodal integration.

\newpage
Table \ref{tabla:sota} presents the performance comparison of several multimodal models for BRCA-SA prediction using clinical, omics, and histopathology image data. The results for all models, except for SAMVAE, are taken directly as reported in the BioFusionNet study. To assess statistical significance, we performed pairwise independent two-sample t-tests based on summary statistics, implemented using the SciPy library \cite{bib43} and applied the HB correction. Each test evaluates whether there is a statistically significant difference between a given model and the best-performing one.

\begin{table}[ht]
\captionsetup{width=0.7\textwidth}
\centering
\begin{tabular}{c|c|c}

\toprule
\multicolumn{1}{c}{Model} & \multicolumn{1}{c}{C-index (Mean ± Std)}  & \multicolumn{1}{c}{p-value HB} \\
\midrule
\textbf{SAMVAE }         &       \textbf{0.699 ± 0.065} & \textbf{0.705} \\
\textbf{MultiSurv }\cite{bib18}     &  \textbf{0.626 ± 0.072} & \textbf{0.057} \\
\textbf{MultiDeepCox-SC }\cite{bib37} &     \textbf{0.604 ± 0.088} & \textbf{0.072} \\
HFBSurv  \cite{bib38}       &        0.536 ± 0.073 & 0.004 \\
PathomicFusion \cite{bib39}   &       0.516 ± 0.080 & 0.005 \\
\textbf{MCAT }   \cite{bib40}        &      \textbf{ 0.700 ± 0.043} & \textbf{0.366}\\
\textbf{TransSurv } \cite{bib41}      &      \textbf{  0.686 ± 0.047} & \textbf{0.207}\\
\textbf{BioFusionNet} \cite{bib34}     &      \textbf{0.768 ± 0.051} & \textbf{1.000 }\\
\bottomrule 
\end{tabular}
\caption[]{Comparison of multimodal models for BRCA-SA prediction using the C-index. Pairwise t-tests were performed to assess significance against the best-performing model, BioFusionNet. We correct p-values using the HB method.}

\label{tabla:sota}
\end{table}  

While BioFusionNet reports the highest mean C-index, the difference compared to our SAMVAE model is not statistically significant (p=0.705), highlighting the competitiveness of our approach. Unlike BioFusionNet, which relies on multiple pretrained models and incurs high computational costs, SAMVAE is trained end-to-end without the need for multiple pretrained components, thereby reducing dependency on the alignment of data modalities with pretrained features. Therefore, we can conclude that SAMVAE is competitive with the state of the art, showing no statistically significant difference with the best-performing model in terms of C-index.

 \subsection{Multimodal Competing Risks}\label{resultscr}

This section presents the results obtained from competing risks cohorts (LGG-CR and BRCA-CR datasets). As detailed in Section \ref{sec2_32}, no multimodal models combining clinical, omics, and imaging modalities, while also modeling continuous time, were found for competing risks. Therefore, we do not compare the results obtained with SAMVAE to the state of the art, and instead present only the ablation study results on the integration of tabular and imaging modalities.

The results of the competing risks analysis for the LGG-CR cohort are presented in Table \ref{tablagrandelgg}, showcasing the performance of our multimodal model across different data modalities. In this case, the omics combination that achieved the best results in intermediate experiments (Appendix \ref{secA3}) consisted of DNAm, CNV, and miRNA data. Regarding imaging, the best performance was obtained when using five patches per patient. These selections were made by optimizing the average CI-IBS score across the two competing risks, ensuring robust performance across both risks simultaneously rather than treating them independently. The CI-IBS metric is defined as the average C-index minus the average IBS across the two risks.  In the LGG-CR setting (see Table \ref{tablagrandelgg}), most combinations of modality appear in bold, indicating statistically significant results (\( p \)-value $>$ 0.01). Additionally, the IBS tends to be higher for Risk 2, likely due to the smaller number of patients who experience this event compared to Risk 1.

In BRCA-CR (see Table \ref{tabla_brca}), the clinical modality alone achieved the highest performance. Regarding omic data, the best-performing combination was CNV + RNA-Seq. By adding histological data (10 patches), the C-index obtained for Risk 1 matched that of the clinical data alone, while achieving a significantly lower IBS. All combinations are highlighted in bold, and similarly to the LGG-CR cohort, we observe that the IBS is higher for Risk 2, which can be explained by the lower number of patients experiencing this event compared to Risk 1.

\begin{table}[ht]
\centering
\begin{subtable}{\textwidth}
\centering
\resizebox{0.9\textwidth}{!}{
\begin{tabular}{|c|c|c|c|c|c|c|c|c|}

\hline
\renewcommand{\arraystretch}{2} % solo afecta esta fila

\makecell{\textbf{Clinical}} &
\makecell{\textbf{DNAm + CNV +}\\\textbf{miRNA}} &
\makecell{\textbf{5 Patches}} &
\makecell{\textbf{Risk}} &
\makecell{\textbf{C-index}} &
\makecell{\textbf{IBS}} & 
\makecell{\textbf{CI} -\\ \textbf{IBS}} &
\makecell{\textbf{p-CI}\\\textbf{HB}} &
\makecell{\textbf{p-IBS}\\\textbf{HB}} \\
\hline
  
\textbullet &   &  & 1 &\textbf{ 0.610 ± 0.044 }& \textbf{0.240 ± 0.046} &      \textbf{ 0.384} &           \textbf{ 1.000 }    &         \textbf{0.221}\\
\hline
\textbullet &   &  & 2 & \textbf{0.616 ± 0.045} & \textbf{0.217 ± 0.018} &      \textbf{ 0.384} &        \textbf{ 1.000}   &      \textbf{ 1.000}     \\
\hline

  & \textbullet &       & 1 & \textbf{0.585 ± 0.028 }& \textbf{0.230 ± 0.047} &   \textbf{  0.352} &              \textbf{0.593}  &         \textbf{0.867} \\
\hline
  & \textbullet &       &  2 & \textbf{0.590 ± 0.032} & \textbf{0.240 ± 0.039} &        \textbf{0.352} &          \textbf{ 0.621}  &        \textbf{ 0.386} \\
\hline

  &   & \textbullet & 1 & \textbf{0.585 ± 0.037} & \textbf{0.227 ± 0.042} &    \textbf{ 0.357} &         \textbf{0.807}  &        \textbf{ 0.952} \\
\hline
  &   & \textbullet & 2 & \textbf{0.595 ± 0.064} & \textbf{0.239 ± 0.042} &      \textbf{0.357} &    \textbf{1.000 }&       \textbf{  0.645} \\
\hline

\textbullet & \textbullet &        &  1 & \textbf{0.580 ± 0.024} & \textbf{0.221 ± 0.039} &     \textbf{0.344} &          \textbf{ 0.266} &       \textbf{  1.000} \\
\hline
\textbullet & \textbullet &        & 2 & \textbf{0.589 ± 0.031} & \textbf{0.260 ± 0.045} &     \textbf{ 0.344} &          \textbf{0.528}  &         \textbf{0.025 }\\
\hline
\textbullet &   & \textbullet    &    1 & \textbf{0.596 ± 0.032} & \textbf{0.204 ± 0.039} &   \textbf{ 0.369} &           \textbf{ 1.000} &        \textbf{1.000 }   \\
\hline
\textbullet &   & \textbullet      &  2 & \textbf{0.587 ± 0.020} & \textbf{0.240 ± 0.021 }&     \textbf{0.369} &           \textbf{ 0.285} &         \textbf{0.028} \\
\hline
  & \textbullet & \textbullet &    1 & \textbf{0.579 ± 0.024} &\textbf{ 0.232 ± 0.057} &      \textbf{0.341} &          \textbf{0.211} &        \textbf{ 0.924} \\
  \hline
  & \textbullet & \textbullet &   2 & 0.591 ± 0.035 & 0.254 ± 0.022&         0.341 &          0.737  &        $<\!1e^{-3}$    \\
\hline
  \textbullet & \textbullet &   \textbullet   &  1 & \textbf{0.577 ± 0.022} &\textbf{ 0.233 ± 0.040} &   \textbf{  0.323} &          \textbf{ 0.146 }&       \textbf{  0.449} \\
  \hline
  \textbullet & \textbullet &   \textbullet   &  2 & 0.570 ± 0.027 & 0.268 ± 0.047 &       0.323 &             0.021 &         0.009 \\
\hline

\end{tabular}
}
\subcaption{Performance metrics for different modality combinations on the LGG-CR.}
\label{tablagrandelgg}
\end{subtable}

\begin{subtable}{\textwidth}
\centering
\resizebox{0.9\textwidth}{!}{ % Increase table size
\begin{tabular}{|c|c|c|c|c|c|c|c|c|c|c|}

\hline
\renewcommand{\arraystretch}{2} % solo afecta esta fila

\makecell{\textbf{Clinical}} &
\makecell{\textbf{CNV + RNAseq}} &
\makecell{\textbf{10 Patches}} &
\makecell{\textbf{Risk}} &
\makecell{\textbf{C-index}} &
\makecell{\textbf{IBS}} &
\makecell{\textbf{CI} -\\ \textbf{IBS}} & 
\makecell{\textbf{p-CI}\\\textbf{HB}} &
\makecell{\textbf{p-IBS}\\\textbf{HB}} \\
\hline
 
\textbullet &   &  & 1 & \textbf{0.664 ± 0.040 }& \textbf{0.199 ± 0.035} &     \textbf{ 0.471} &      \textbf{ 1.000 }  &         \textbf{0.806} \\
\hline
\textbullet &   &  &  2 & \textbf{0.682 ± 0.042} & \textbf{0.205 ± 0.033} &        \textbf{ 0.471} &    \textbf{ 1.000 }    &         \textbf{1.000} \\
\hline

  & \textbullet &       & 1 & \textbf{0.618 ± 0.055} & \textbf{0.195 ± 0.028} &     \textbf{  0.433} &             \textbf{ 0.296} &         \textbf{1.000} \\
\hline
  & \textbullet &       & 2 & \textbf{0.632 ± 0.053} & \textbf{0.190 ± 0.031} &    \textbf{0.433} &          \textbf{0.074} &        \textbf{1.000 }   \\
\hline
  &   & \textbullet &  1 & \textbf{0.638 ± 0.037} & \textbf{0.206 ± 0.030} &      \textbf{ 0.433} &         \textbf{1.000} &        \textbf{ 0.190}   \\
\hline
  &   & \textbullet &  2 &\textbf{ 0.635 ± 0.039} & \textbf{0.200 ± 0.022} &    \textbf{ 0.433} &        \textbf{ 0.035}  &         \textbf{1.000}  \\
\hline

\textbullet & \textbullet &        &     1 &\textbf{ 0.659 ± 0.055} & \textbf{0.177 ± 0.036} &      \textbf{0.456} &       \textbf{1.000}    &   \textbf{ 1.000 }   \\
\hline
\textbullet & \textbullet &        &  2 & \textbf{0.632 ± 0.060} &\textbf{ 0.200 ± 0.035} &       \textbf{0.456} &             \textbf{0.112} &      \textbf{   1.000}  \\
\hline
\textbullet &   & \textbullet    &    1 & \textbf{0.657 ± 0.060} &\textbf{ 0.196 ± 0.023} &    \textbf{0.446} &           \textbf{ 1.000}     &        \textbf{ 0.694} \\
\hline
\textbullet &   & \textbullet      &  2 & \textbf{0.639 ± 0.026} & \textbf{0.207 ± 0.021} &       \textbf{0.446} &         \textbf{ 0.024} &        \textbf{ 0.667} \\  
\hline
  & \textbullet & \textbullet &      1 & \textbf{0.641 ± 0.038 }& \textbf{0.204 ± 0.033} &     \textbf{0.433} &           \textbf{ 1.000 }&        \textbf{ 0.357} \\
  \hline
  & \textbullet & \textbullet &   2 &\textbf{ 0.639 ± 0.044} & \textbf{0.210 ± 0.042} &    \textbf{0.433} &              \textbf{0.088} &       \textbf{  1.000} \\
\hline
  \textbullet & \textbullet &   \textbullet   &  1 & \textbf{0.666 ± 0.066} & \textbf{0.183 ± 0.037} &      \textbf{0.458} &       \textbf{1.000  }   &        \textbf{1.000}  \\
  \hline
  \textbullet & \textbullet &   \textbullet   &  2 & \textbf{0.637 ± 0.031} & \textbf{0.205 ± 0.032} &   \textbf{  0.458} &         \textbf{ 0.024 }&        \textbf{1.000} \\
 
\hline
\end{tabular}
}
\subcaption{Performance metrics for different modality combinations on the BRCA-CR.}
\label{tabla_brca}
\end{subtable}

\caption{Performance metrics for different modality combinations on the LGG-CR and BRCA-CR datasets. The tables report the C-index, IBS, and their difference (CI - IBS). P-values (corrected using HB) are computed by comparing each combination against the best-performing one in terms of C-index and IBS (for each risk), respectively. Bold values indicate combinations whose performance is not significantly different ($p > 0.01$) from the best one.}
\label{tab:grandesubtabla}
\end{table}   
 
\begin{figure}[ht]
\centering

\begin{subfigure}[b]{0.8\textwidth}
    \centering
    \includegraphics[width=\textwidth]{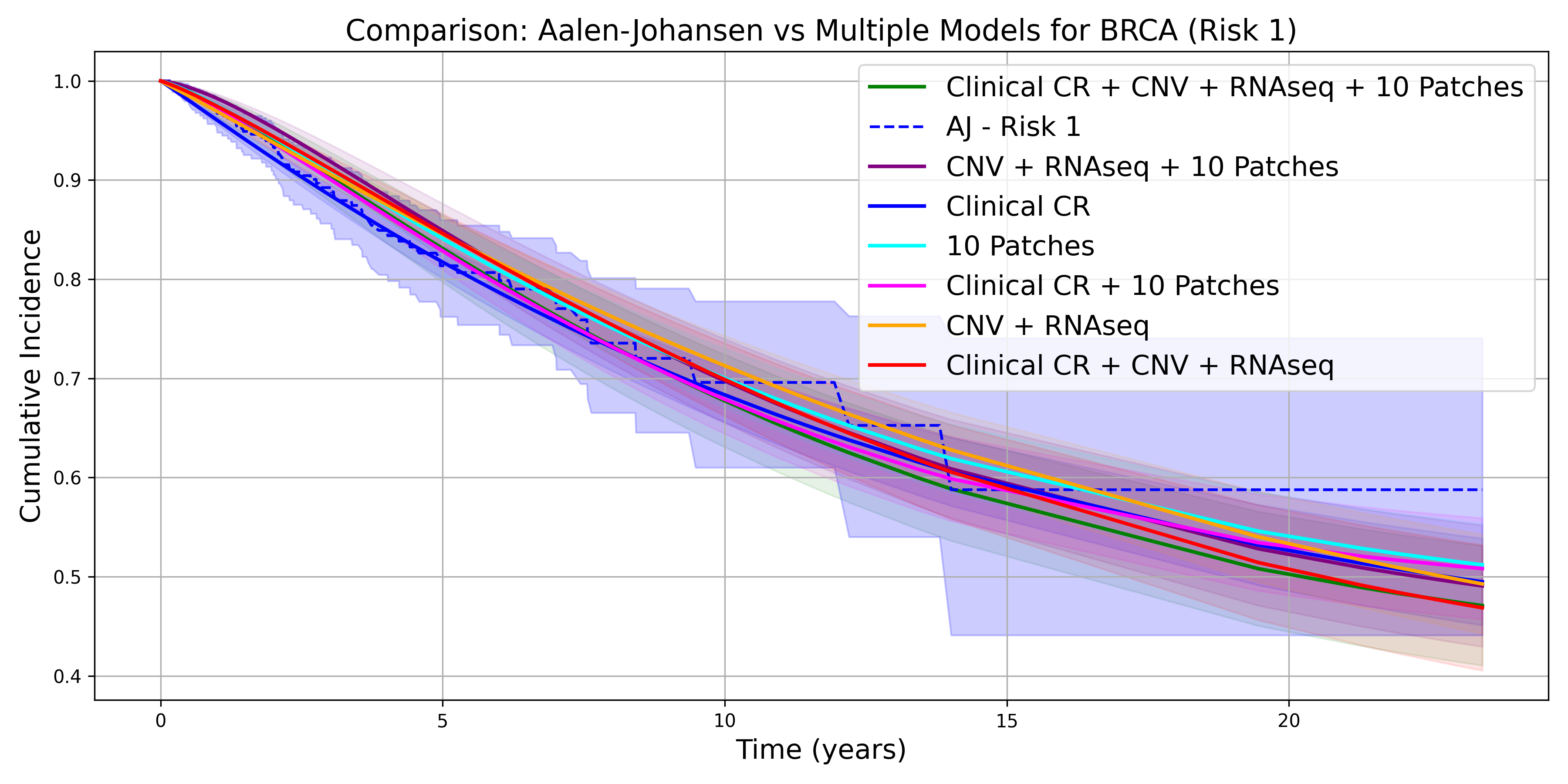}
    \subcaption{Average predicted CIFs for Risk 1 in the BRCA-CR dataset.}
    \label{risk1_brca}
\end{subfigure}

\begin{subfigure}[b]{0.8\textwidth}
    \centering
    \includegraphics[width=\textwidth]{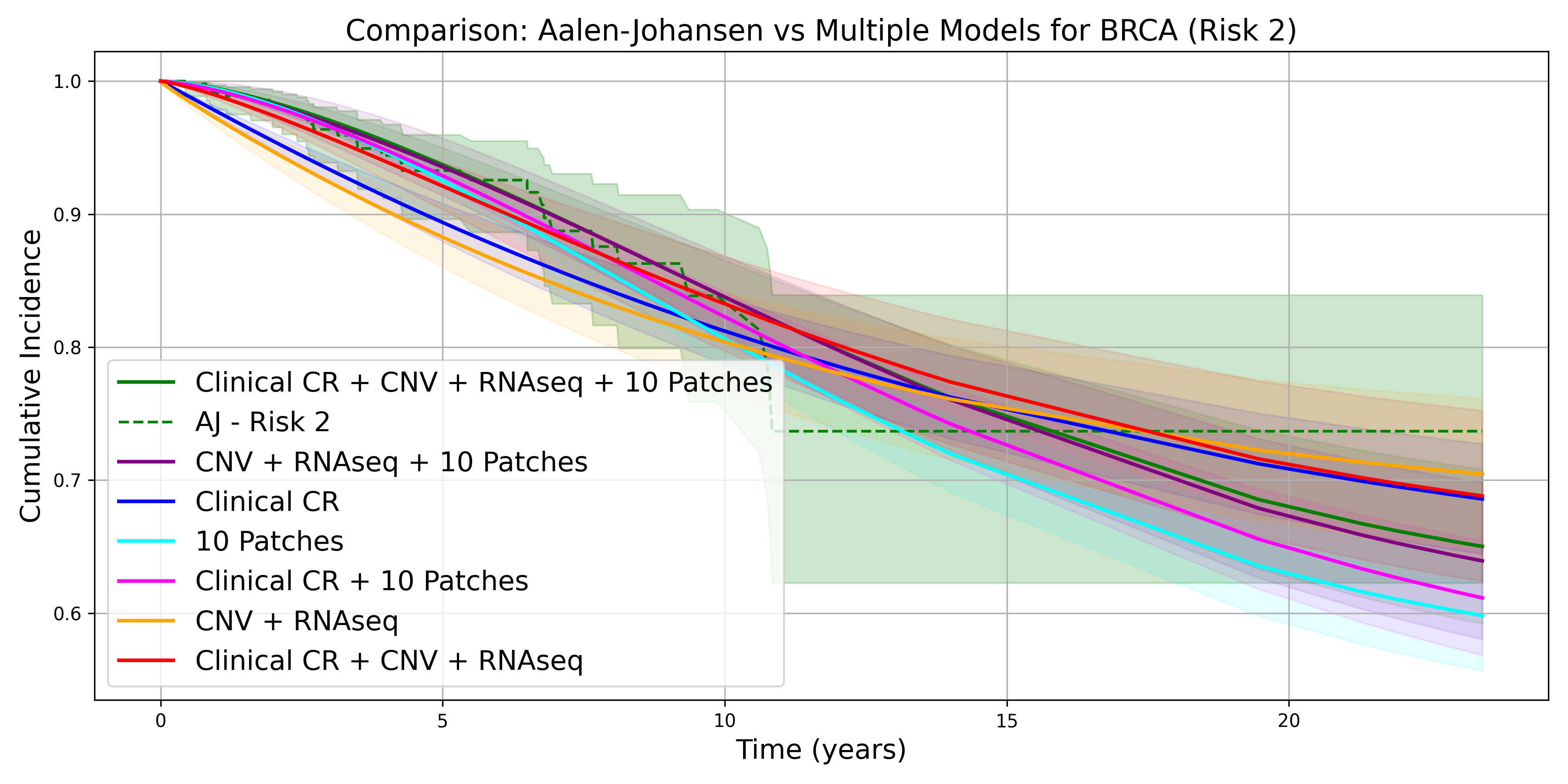}
    \subcaption{Average predicted CIFs for Risk 2 in the BRCA-CR dataset.}
    \label{risk2_brca}
\end{subfigure}

\caption{Average predicted CIFs for the two competing risks in the BRCA-CR dataset. The predictions are compared against the AJ estimator, which properly handles competing risks. Each colored line corresponds to a different combination of input modalities used by the model. The close overlap with the AJ curves indicates accurate population-level estimation.}
\label{fig:brca_cr_cif}
\end{figure}

\begin{figure}[ht]
\centering

\begin{subfigure}[b]{0.8\textwidth}
    \centering
    \includegraphics[width=\textwidth]{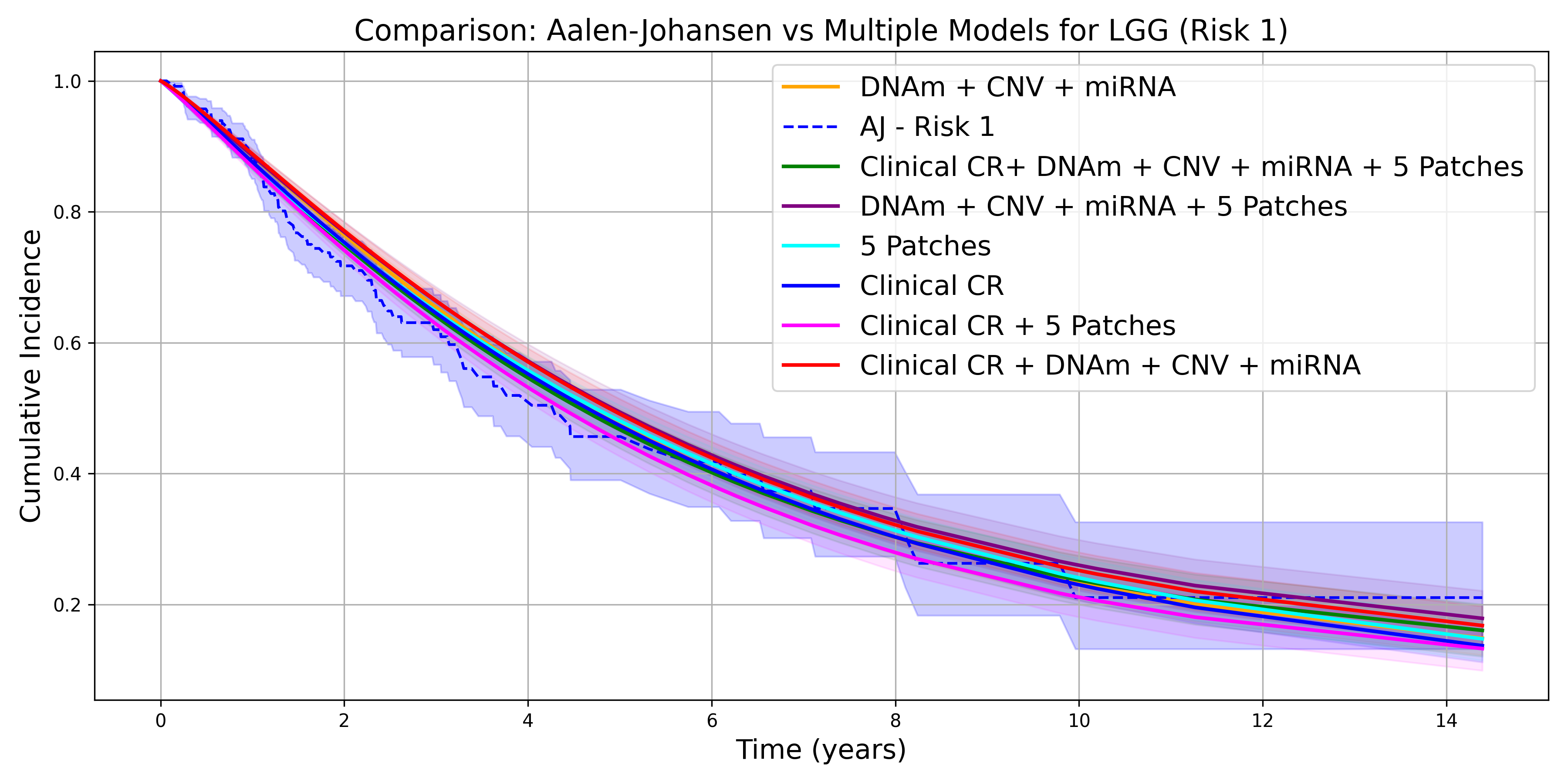}
    \subcaption{Average predicted CIFs for Risk 1 in the LGG-CR dataset.}
    \label{risk1_lgg}
\end{subfigure}

\begin{subfigure}[b]{0.8\textwidth}
    \centering
    \includegraphics[width=\textwidth]{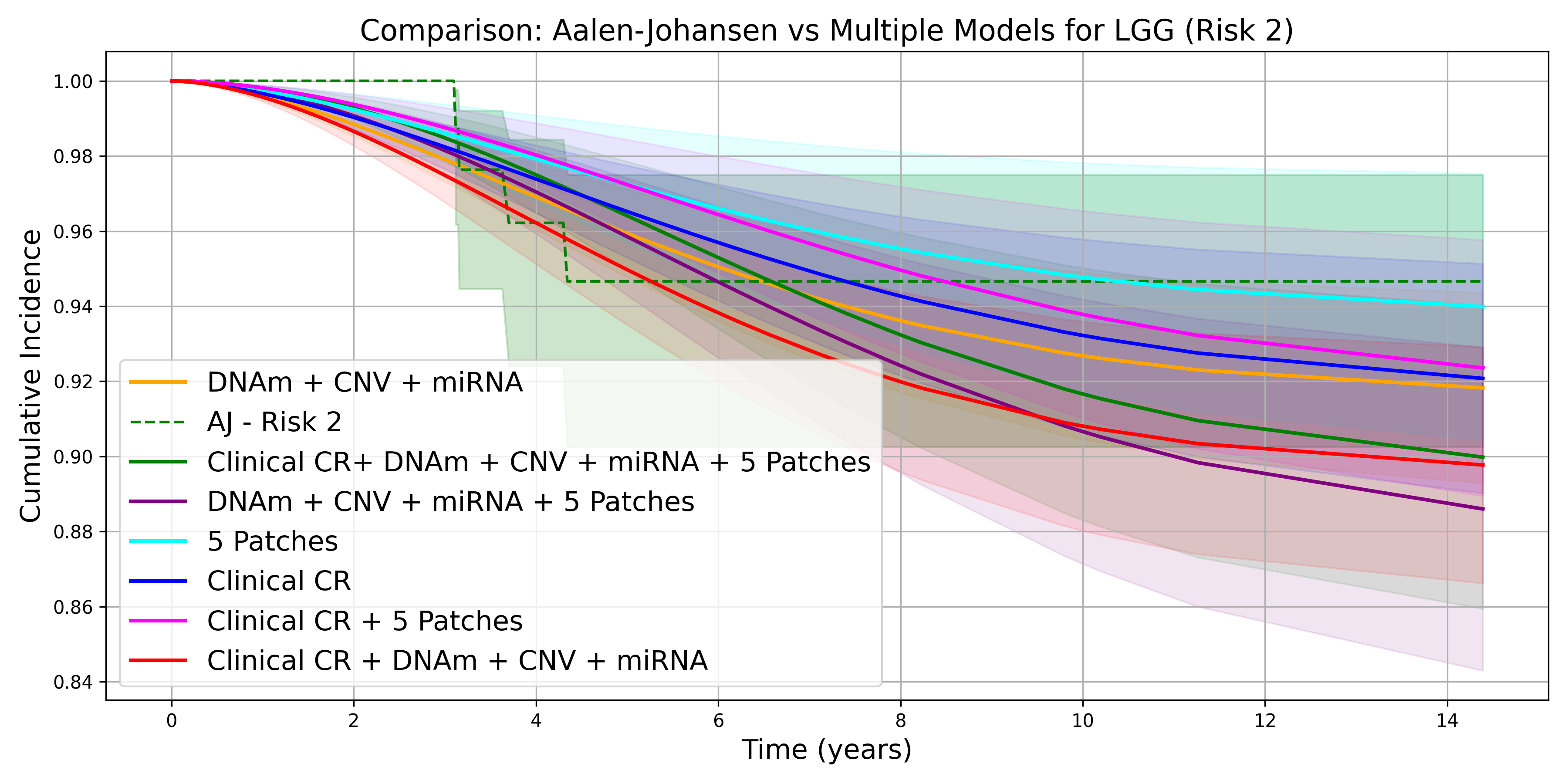}
    \subcaption{Average predicted CIFs for Risk 2 in the LGG-CR dataset.}
    \label{risk2_lgg}
\end{subfigure}

\caption{The curves represent the model's average predicted CIFs across patients, estimated using the AJ estimator, which properly accounts for competing risks. Each colored line corresponds to a different combination of input modalities used by the model. The close overlap with the AJ curves indicates accurate population-level estimation.}
\label{lgg_cr_cif}
\end{figure}

To better visualize this, we illustrate the CIFs obtained with each combination of modalities for each competing risk and the Aalen-Johansen (AJ) estimator \cite{bib22}, for the BRCA-CR cohort (Figures \ref{risk1_brca} and \ref{risk2_brca}) and the LGG-CR cohort (Figures \ref{risk1_lgg} and \ref{risk2_lgg}).

Unlike the KM estimator, which treats competing events as censored and thus tends to overestimate the probability of each individual event, the AJ estimator properly accounts for the presence of competing risks \cite{bib20}, providing a more realistic and accurate estimate of the event-specific survival probability. Figures \ref{risk1_brca} and \ref{risk2_brca} show that the predicted CIFs produced by our models for BRCA-competing risks exhibit strong overlap with the AJ estimates, indicating that, at the population level, the models accurately capture the dynamics of cumulative incidence. The only noticeable deviation appears for Risk 2 in the omics-only configuration, which slightly overestimates the incidence. In contrast, the rest of the modality combinations align closely with the AJ curve. Additionally, we observe that the confidence intervals around the AJ estimator widen considerably after year 10, reflecting increasing uncertainty due to the reduced number of observed events over time. This highlights the limitations of empirical estimators in the presence of data sparsity and reinforces the relevance of model-based approaches under such conditions.

Figures \ref{risk1_lgg} and \ref{risk2_lgg} show that, for the LGG-CR cohort, the predicted CIFs align closely with the AJ estimates across all modality configurations. For Risk 1, the curves exhibit a strong and consistent overlap with the AJ estimator. In the case of Risk 2, however, we observe considerably wider confidence intervals around the AJ curve, particularly after year 5. This is a consequence of the substantially smaller number of events associated with Risk 2 compared to Risk 1, which increases statistical uncertainty.

Overall, these findings demonstrate a successful integration of multimodal data for competing risks modeling. The fact that multimodal combinations do not underperform compared to clinical data alone, as consistently observed in Tables \ref{tablagrandelgg} and \ref{tabla_brca}, reinforces the complementary value of omic and histological information. 
 
To the best of our knowledge, there are currently no existing multimodal deep learning frameworks that address competing risks modeling in continuous time (rather than just discrete time) to specific events, while simultaneously integrating clinical, omics, and imaging modalities, as detailed in Section \ref{sec2_32}. As a result, direct comparisons with state-of-the-art models are not feasible.

\subsection{Personalized medicine}
One of the key advantages of the proposed approach is its ability to generate personalized survival curves for each patient, accounting for multiple plausible scenarios by sampling from the latent space. This allows the model to capture individual uncertainty and generate distributions over survival times, offering a richer and more informative prognosis than point estimates. These individualized trajectories can inform clinical decisions such as treatment intensity, follow-up frequency, or eligibility for clinical trials.

\begin{figure}[ht]
\centering
\includegraphics[width=0.8\textwidth]{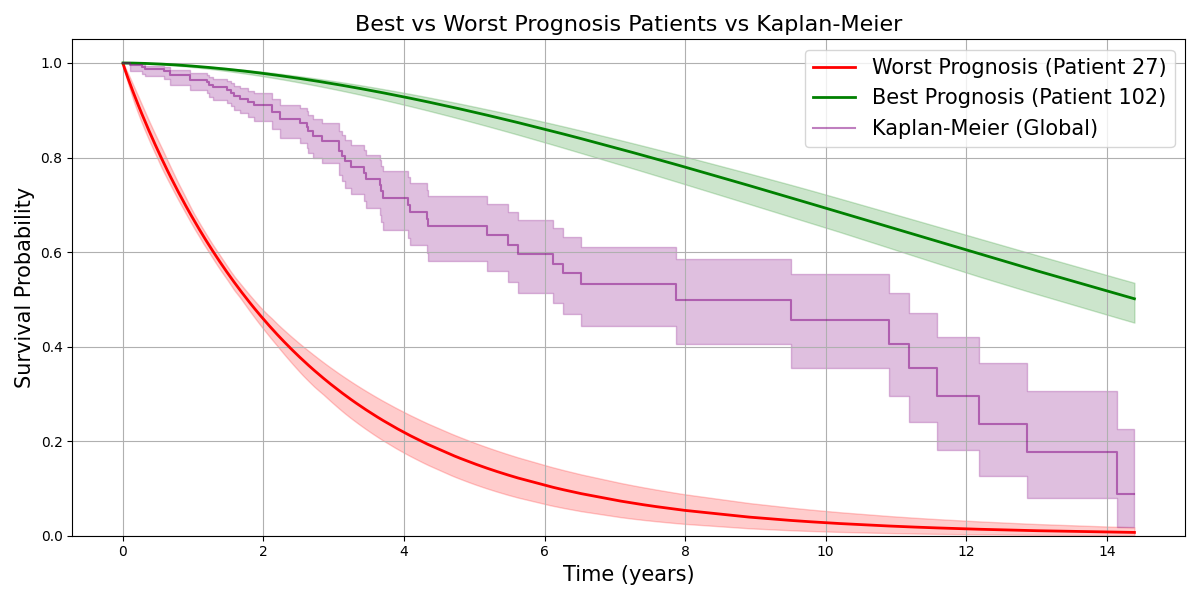}
\caption{Model-predicted survival for patients with best and worst prognoses, alongside the cohort-level KM curve. Shaded areas denote 90\% confidence intervals.  For the patient with the worst prognosis, clinical and selected omic data were reviewed, revealing an 80-year-old individual with apparent TP53 activation. Such patient-level predictions can be obtained for any test case, enabling clinicians to provide personalized prognosis. An interactive notebook that allows users to explore the clinical features associated with each patient's survival curve is available at \cite{bib42}}
\label{fig:best_worst_patient}
\end{figure}  

To illustrate the personalized nature and practical utility of our approach, Figure \ref{fig:best_worst_patient} presents a comparison of the survival curves predicted by our model for the patients with the best and worst prognoses in the test set for LGG, alongside the empirical KM estimator for the entire cohort. For each selected patient, we generated 100 survival trajectories by sampling from the posterior distribution of the latent space. The mean survival curve is shown with a 90\% confidence band (5th–95th percentiles). The patient with the poorest prognosis shows a steep decline in survival probability shortly after diagnosis, while the best prognosis patient maintains a high probability of survival throughout the follow-up period.

The project repository includes an interactive HTML visualization that allows users to explore the clinical features associated with each patient's survival curve for LGG (see Appendix \ref{sece2}, available at \href{https://albagarridolopezz.github.io/SAMVAE/best_vs_worst_prognosis_interactive.html}{Best vs Worst Prognosis Patients for LGG}). For instance, in Figure \ref{fig:best_worst_patient}, the green curve corresponds to a 35-year-old patient with a favorable prognosis, while the red curve represents an 80-year-old patient with a much poorer expected outcome. In addition to age, other relevant features such as prior treatments or molecular characteristics can also be examined, enabling a more comprehensive and transparent clinical interpretation of the model’s predictions.

For the patient with the worst prognosis, we observed that in the CNV omics modality, genes such as TP53 appeared amplified. This gene is a well-known prognostic factor in lower-grade glioma \cite{bib25}. This illustrates how our model enables clinically meaningful interpretation: the predicted survival curve becomes more than just a number when a clinician can relate it to specific omic patterns. Such patient-level predictions can be obtained for any test case, enabling clinicians to provide personalized prognosis.

\section{Conclusion}\label{sec13}
 In this work, we introduced SAMVAE, a novel deep learning architecture designed to address the challenges of survival prediction in oncology through effective multimodal data integration while modeling continuous time. By combining clinical information, molecular profiles, and histopathological features into a shared latent space, SAMVAE achieves robust and accurate survival prediction while preserving modality-specific representations. In addition, this unified probabilistic framework enables the generation of personalized and clinically interpretable survival curves.

It is important to emphasize that our approach intentionally avoided extensive preprocessing and complex feature engineering, thereby underscoring the considerable untapped potential of omics and imaging data. These findings support continued investment in multimodal survival models, particularly as data availability and computational methodologies continue to advance.

Notably, both the omic and histological pipelines were kept intentionally simple and reproducible, without aggressive feature selection or hyperparameter tuning. As such, the observed improvements from modality integration likely represent conservative estimates of the true potential of multimodal fusion. These results underscore the value of integrating diverse data modalities in survival analysis, demonstrating that even with minimal preprocessing, multimodal approaches can offer complementary information beyond conventional clinical variables. Future work should focus on more advanced integration techniques and deeper feature extraction from non-clinical sources. Moreover, understanding what features are relevant for each specific cancer type may further enhance this process, enabling more effective tailoring of the SAMVAE framework to concrete cancer settings.

 Although the proposed approach can also handle discrete time distributions, provided their log-likelihood is differentiable, we focus exclusively on a continuous Weibull distribution, as it is widely used in survival analysis \cite{bib70} and does not assume proportional hazards. Furthermore, unlike models based exclusively on discrete time, SAMVAE does not suffer from the limitation of requiring large patient cohorts to achieve reliable performance.

 SAMVAE achieves competitive performance in the multimodal survival analysis setting compared to using clinical data alone, showing comparable results in terms of the C-index and consistent improvements in the IBS. The superior calibration reflected in the IBS highlights the model’s strength in estimating reliable survival probabilities. This is especially relevant in clinical settings where accurate probabilistic predictions over time are more informative than mere ranking of patients. 

  As expected, clinical variables — particularly age — dominate survival prediction, which aligns with existing clinical knowledge and confirms the relevance of classical prognostic factors \cite{bib23} (see clinical explainability in Appendix \ref{xai2} and \ref{xai1} for further details). However, despite the strong predictive power of clinical variables alone, our study demonstrates that integrating additional omic and imaging modalities leads to well-calibrated models. The consistent reduction in the IBS when combining modalities suggests that non-clinical data contributes meaningful complementary information. 
 
Our model enables individualized survival prediction by leveraging multimodal data to generate patient-specific survival curves with uncertainty estimates. Unlike traditional survival models that rely on fixed covariate effects, our approach captures complex, nonlinear interactions across clinical and molecular features. This allows us to stratify patients beyond standard clinical risk groups and to identify those who may benefit from intensified follow-up or targeted interventions. As shown in Figure \ref{fig:best_worst_patient}, this capability opens the door to data-driven precision medicine, where prognosis and clinical decisions are tailored to the unique characteristics of each patient. 

In the case of competing risks settings, our results indicate a successful multimodal integration, as SAMVAE achieves competitive performance compared to models trained solely on clinical data. This demonstrates that incorporating omics and histological information contributes meaningful, complementary insights beyond standard clinical variables. To the best of our knowledge, this is the first multimodal deep learning framework that explicitly addresses competing risks by modeling time to a specific event in continuous domains. This marks a significant step forward in personalized survival prediction across heterogeneous data sources. Given the lack of existing multimodal benchmarks in this area that include clinical, omics, and imaging data, direct comparisons with state-of-the-art models are not currently feasible.

In conclusion, SAMVAE provides a successful framework for multimodal integration, effectively combining omics, imaging, and clinical data into a shared latent space. In addition, our SAMVAE model achieves competitive performance with the current state-of-the-art while relying on a significantly simpler and more unified architecture. Thanks to its end-to-end design, SAMVAE eliminates the need for complex training pipelines or multiple specialized and pretrained modules. In contrast to black-box models, SAMVAE's probabilistic outputs and interpretable structure facilitate clinical insight, allowing practitioners to assess how different modalities influence risk over time. This interpretability, combined with its ability to handle competing risks, a setting not addressed by existing multimodal survival models, positions SAMVAE as a powerful and practical tool for multimodal survival analysis in biomedical research.

Several directions remain open as future work for improving the proposed SAMVAE framework:
\begin{itemize}
    \item Modality-aware weighting loss: In its current form, SAMVAE treats all input modalities uniformly. Introducing modality-specific weighting mechanisms, as suggested in \cite{bib11}, such as those inspired by the weighted Cox hazard approach in \cite{bib34}. It could allow the model to dynamically prioritize more informative data sources during training, potentially improving predictive performance and interpretability.
    \item Although omics and imaging data were preprocessed to ensure basic compatibility, we did not conduct a systematic evaluation of how different preprocessing pipelines affect downstream performance. Future work will investigate the impact of alternative normalization, feature selection, and dimensionality reduction techniques on further enhancing the robustness and generalization of multimodal integration across diverse biomedical datasets.
    \item Another direction is the development of methods for generating synthetic multimodal patient profiles, which could revolutionize how we address data limitations and resource constraints in clinical practice. A key question for future exploration is whether missing modalities, such as imaging or omics data, can be accurately generated from available modalities (e.g., clinical and genomic data) while maintaining reliable survival predictions. This capability could enable more equitable healthcare by providing robust prognoses for patients with fewer available tests, addressing disparities in resource-limited settings. 
 
\end{itemize}

%\backmatter
 
\section*{Declarations}
 
\paragraph{Funding} This project is supported by the Innovative Health Initiative Joint Undertaking (IHI JU) under grant agreement No 101172872 (SYNTHIA). The JU receives support from the European Union's Horizon Europe research and innovation programme, COCIR, EFPIA, Europa Bío, MedTech Europe, Vaccines Europe and DNV. The UK consortium partner, The National Institute for Health and Care Excellence (NICE) is supported by UKRI Grant 10132181. Funded by the European Union, the private members, and those contributing partners of the IHI JU. Views and opinions expressed are however those of the author(s) only and do not necessarily reflect those of the aforementioned parties. Neither of the aforementioned parties can be held responsible for them. 
\paragraph{Competing interests} The authors declare no competing interests.

\paragraph{\textbf{Consent for publication}} Not applicable.
 
\paragraph{\textbf{Data availability}}  The data used in this study were obtained from The Cancer Genome Atlas (TCGA), a publicly accessible repository available through the GDC Data Portal (https://portal.gdc.cancer.gov/). A step-by-step guide detailing the data download process and applied filters can be found in the following repository:  https://github.com/AlbaGarridoLopezz/SAMVAE (accessed on 7 July 2025).
\paragraph{\textbf{Ethics approval and consent to participate}} According to the TCGA data usage policies, all data used are publicly available and fully anonymized. Therefore, no ethical approval was required for their use in this research.
\paragraph{\textbf{Materials availability}}   Not applicable.
\paragraph{\textbf{Code availability}} The code used in this research is publicly available and can be found in the following repository: https://github.com/AlbaGarridoLopezz/SAMVAE (accessed on 7 July 2025).
\paragraph{\textbf{Author contribution}} All authors contributed to the conceptualization of the study. A.G., A.A., and P.A.A. designed the general architecture. A.G. and P.A.A. were responsible for software implementation and validation. Data curation was performed by A.G. Formal analysis was conducted by J.P. and S.Z., who also oversaw project administration and supervision together with A.A. A.G. wrote the original draft of the manuscript. All authors reviewed and approved the final version of the manuscript.

\noindent 

\bigskip
\begin{flushleft}%

%\bigskip\noindent
%Springer journals and proceedings: \url{https://www.springer.com/gp/editorial-policies}

%\bigskip\noindent
%Nature Portfolio journals: \url{https://www.nature.com/nature-research/editorial-policies}

%\bigskip\noindent
%\textit{Scientific Reports}: \url{https://www.nature.com/srep/journal-policies/editorial-policies}

%\bigskip\noindent
%BMC journals: \url{https://www.biomedcentral.com/getpublished/editorial-policies}
\end{flushleft}

\begin{appendices}

\section{Notation}\label{secA2}

We summarize below the notation used throughout the paper:

\textbf{General Variables}

\begin{itemize}
    
\item $N$: Total number of samples (patients) in the dataset.
\item $K$: Number of possible event types (e.g., different causes of failure in competing risks). 
\item $L$: Number of covariates (features) per sample.
\item $M$: Number of input modalities (e.g., clinical, omics, images).
\end{itemize}
 \textbf{ Data Structure}

Each sample is represented by a triplet:

$$
(\covariablesi, \timei,\eventi) \quad \text{for } i = 1, \dots, N
$$

where:
\begin{itemize}
\item $\covariablesi^{\langle m \rangle} \in \mathbb{R}^{L_m}$: Covariate vector of modality $m$ for the $i$-th individual, where $L_m$ denotes the number of features associated with modality $m$. Each modality may have a different dimensionality.

\item $\timei \in \mathbb{R}_{\geq 0}$: Observed time (event or censoring).
\item  $\eventi \in \{0, 1\}$ for standard survival analysis, or $\eventi \in \{0, 1, \dots, K\}$ for competing risks.

  \item $\eventi = 0$: Right-censored observation (no event observed during follow-up).
  \item  $\eventi = k \in \{1, \dots, K\}$: Event of type $k$ occurred.
\end{itemize}

 \textbf{Model Components}
\begin{itemize}
\item $\phi$: Encoder functions for each modality.
\item $\theta$: Decoder for covariates.
\item $\psi$: Decoder for survival time(s).
\item $\omega$: Classifier for event type (if competing risks).
\item $z^{\langle m \rangle}$: Latent representation of modality $m$
\end{itemize}

 \textbf{Notes}
\begin{itemize}
    \item In standard survival analysis, only a single event type is considered ($K = 1$), and $\eventi \in \{0, 1\}$.
\item In competing risks settings, multiple types of events are modeled, and the event indicator $\eventi$ explicitly encodes the type of event or censoring. 

\end{itemize}

\section{Parameters Tested and Optimals}\label{secA1}

This section summarizes the hyperparameter values explored and the optimal configurations selected for each modality and task (survival analysis and competing risks). Table \ref{values_tested} presents the range of latent dimensions and hidden layer sizes tested during model development. For the omic and imaging modalities, hyperparameter tuning was performed in the presence of clinical data, ensuring that the selected configurations reflect the most effective multimodal integration scenarios. Table \ref{tableopt} reports the optimal parameter combinations for each modality, dataset (LGG and BRCA), and task (standard survival analysis and competing risks modeling).

\begin{table}[ht]
\centering
\begin{tabular}{|c|c|c|}
\hline
\textbf{Mode}           & \textbf{Hyperparameter}   & \textbf{Values Tested}                                \\ \hline
Clinical               & Latent Dimension          & 5                                                   \\ 
                       & Hidden Size               & 5, 10, 25, 50, 75, 100, 500                                                                          \\ \hline
Omics  
                       & Latent Dimension          & 5, 50, 500                                          \\ 
                       & Hidden Size               & 5, 50, 500                                          \\  \hline
Images                   
                       & Latent Dimension          & 5, 50, 500                                          \\ 
                       & Hidden Size               & \{8, 16, 32\}, \{16, 32, 64\}, \{32, 64, 128\}                                                  \\ \hline
\end{tabular}
\caption{Hyperparameters tested for each modality.}
\label{values_tested}
\end{table}

 \begin{table}[ht]
\centering
\scriptsize

\begin{tabular}{|l|l|c|c|c|c|}
\hline
\textbf{Dataset} & \textbf{Modality} & \multicolumn{2}{c|}{\textbf{Survival Analysis}} & \multicolumn{2}{c|}{\textbf{Competing Risks}} \\
 
 & & Latent Dimension & Hidden Size & Latent Dimension & Hidden Size \\
\hline
\multirow{7}{*}{\textbf{LGG}} 
& Clinical       & 10 & 10             & 5  & 75 \\
& DNAm     & 5  & 50             & 5  & 50 \\
& CNV     & 5  & 500            & 5  & 500 \\
& miRNA    & 5  & 5              & 5  & 50 \\
& RNAseq   & 5  & 500            & 5  & 50 \\
& Patches   & 5  & [8, 16, 32]     & 5  & [16, 32, 64] \\
 \hline
\multirow{7}{*}{\textbf{BRCA}} 
& Clinical       & 10 & 100            & 10 & 500 \\
& DNAm      & 5  & 5              & 5  & 5 \\
& CNV      & 5  & 5              & 5  & 50 \\
& miRNA    & 5  & 5              & 5  & 5 \\
& RNAseq   & 5  & 500            & 50 & 50 \\
& Patches   & 5  & [32, 64, 128]   & 50 & [16, 32, 64] \\
 \hline
\end{tabular}
\caption{Optimal parameters (Latent Dimension and Hidden Size) for each modality, dataset, and scenario}
\label{tableopt}
\end{table}

\section{Intermediate Results}\label{secA3}

 \subsection{Hyperparameter Tuning}
As an illustrative example, we include in this appendix a subset of the intermediate tables generated during the hyperparameter optimization process for survival analysis and competing risks scenarios. All omics and imaging modalities were optimized in conjunction with clinical data, which served as the baseline for multimodal integration. Additional hyperparameter tuning results and complete tables for other modality combinations and scenarios (BRCA-SA, BRCA-CR, LGG-SA and LGG-CR) are available in the public repository \cite{bib42}.

 Specifically, we present one of the tables used to select the optimal latent dimension for the LGG-SA prediction task, based on omics data (DNAm) combined with clinical variables (see Table \ref{table:ADN}). In this table, the first value of the latent dimension and the hidden size correspond to the optimal value previously determined for the clinical data alone and is therefore fixed across all combinations. The second value represents the latent dimension tested for the DNAm modality. Discarding all combinations with any corrected \( p \)-value below 0.01, the configuration with the best C-index–IBS trade-off among the remaining options is shaded in a different colour. In this case, the optimal combination for DNAm corresponds to a latent space dimension of 5 and a hidden size of 50.

\begin{table}[ht]
\centering
\setlength{\tabcolsep}{5pt}
\renewcommand{\arraystretch}{1.0}
\resizebox{0.8\textwidth}{!}{ 
\begin{tabular}{|>{  }c|>{ }c|c|c|c|c|c|c|c|c|c|c|}
\hline
 \textbf{Latent}    & \textbf{Hidden}    & \textbf{C-index}       & \textbf{IBS}           & \textbf{CI - IBS}      & \textbf{p-CI HB} &   \textbf{p-IBS HB} \\
\hline
  \textbf{\textnormal{[10, 5]}}   &  \textbf{\textnormal{[10, 5]}}   & \textbf{0.685 ± 0.078} & \textbf{0.204 ± 0.027} & \textbf{0.481 ± 0.093 }&                \textbf{ 0.318} &              \textbf{0.096} \\
\rowcolor[HTML]{D8A9C4} \textbf{ \textnormal{[10, 5]}}   & \textbf{ \textnormal{[10, 50]}}  &\textbf{ 0.717 ± 0.066} & \textbf{0.183 ± 0.029} & \textbf{0.533 ± 0.086} &              \textbf{     0.996} &              \textbf{0.500 }  \\
\textbf{\textnormal{[10, 5]} }  &  \textbf{\textnormal{[10, 500]}} &\textbf{ 0.717 ± 0.051} & \textbf{0.192 ± 0.025} &\textbf{ 0.525 ± 0.057} &                   \textbf{0.996} &              \textbf{0.404} \\
\textnormal{[10, 50]}   &  \textnormal{[10, 5]}  & 0.610 ± 0.053 & 0.217 ± 0.028 & 0.393 ± 0.067 &                  0.000     &              0.014 \\
\textnormal{[10, 50]}   &  \textnormal{[10, 50]}   & 0.603 ± 0.069 & 0.223 ± 0.019 & 0.380 ± 0.067 &                 0.000     &              0.000     \\
 \textbf{\textnormal{[10, 50]}}   & \textbf{ \textnormal{[10, 500]}}  &\textbf{ 0.664 ± 0.061 }& \textbf{0.205 ± 0.025} & \textbf{0.459 ± 0.062 }&                      \textbf{0.036} &            \textbf{  0.096} \\
\textnormal{[10, 500]}   &  \textnormal{[10, 5]}    & 0.556 ± 0.038 & 0.207 ± 0.019 & 0.348 ± 0.047 &                   0.000     &              0.042 \\
\textnormal{[10, 500]}   &  \textnormal{[10, 50]}  & 0.601 ± 0.065 & 0.217 ± 0.038 & 0.384 ± 0.082 &                     0.000     &              0.042 \\
 \textnormal{[10, 500]}   &  \textnormal{[10, 500]} & 0.589 ± 0.073 & 0.253 ± 0.054 & 0.337 ± 0.071 &                 0.000    &              0.000     \\
\hline
\end{tabular}}

\caption{Parameter Combination for DNAm with Clinical Data. The \( p \)-values corrected with HB that exceed 0.01 are in bold. Discarding combinations with any \( p \)-value below 0.05, the one with the best CI-IBS is shaded in a different colour. p-CI HB and p-IBS HB represent the \( p \)-values for the C-Index and IBS corrected with HB, respectively.}
\label{table:ADN}
\end{table}  

  \subsection{Intermediate Configurations}
Additionally, once the optimal hyperparameters were selected for each individual modality (including images), we systematically evaluated all possible combinations of omic modalities and experimented with different numbers of image patches per patient to determine the best configuration for image-based models. As examples, we include the table summarizing the performance of all omic modality combinations for the BRCA-SA task (Table \ref{omicbrca}), and another showing the image patch selection process for LGG-CR (Table \ref{wsilgg}).  For instance, in the BRCA-SA case, all combinations except one are highlighted in bold, as they surpass the statistical threshold of  \( p \)-value $>$ 0.01. Notably, the CNV modality achieves the best CI-IBS score among all, and is therefore shaded differently to emphasize its superior performance.

\begin{table}[ht]
\centering
\resizebox{0.9 \textwidth}{!}{ % Increase table size
\begin{tabular}{|c|c|c|c|c|c|c|c|c|c|c|}
\rowcolor[gray]{1}
\hline
\textbf{Clinical} & \textbf{DNAm} & \textbf{CNV} & \textbf{miRNA} & \textbf{RNAseq}  & \textbf{C-index} & \textbf{IBS} & \textbf{CI - IBS} &  \textbf{p-CI HB} & \textbf{p-IBS HB} \\
\hline
\textbullet &\textbullet &   &   &   & \textbf{0.661 ± 0.071} &\textbf{ 0.200 ± 0.042} & \textbf{0.461 ± 0.086} &       \textbf{ 1.000  }   &         \textbf{0.768}  \\
\hline
\rowcolor[HTML]{D8A9C4} \textbullet &  & \textbullet &   &   &  \textbf{0.684 ± 0.053} & \textbf{0.175 ± 0.034} & \textbf{0.509 ± 0.069} &        \textbf{  1.000  }    &       \textbf{ 1.000}      \\
\hline
\textbullet &  &   & \textbullet &  & \textbf{0.687 ± 0.072} &\textbf{ 0.207 ± 0.039} &\textbf{ 0.480 ± 0.084} &              \textbf{1.000}      &         \textbf{0.219} \\
\hline
\textbullet &  &   &   & \textbullet  & \textbf{0.654 ± 0.051} & \textbf{0.186 ± 0.020} & \textbf{0.468 ± 0.043} &         \textbf{1.000}        &   \textbf{   1.000   }   \\
\hline
\textbullet & \textbullet & \textbullet &   &     & \textbf{0.641 ± 0.041} & \textbf{0.208 ± 0.039} &\textbf{ 0.432 ± 0.047} &            \textbf{ 0.350 } &         \textbf{0.191} \\ 
\hline
\textbullet & \textbullet &   & \textbullet &    & \textbf{0.604 ± 0.050} & \textbf{0.197 ± 0.030} & \textbf{0.407 ± 0.070} &            \textbf{ 0.012} &      \textbf{   0.628} \\
\hline
\textbullet & \textbullet &   &   & \textbullet       &\textbf{ 0.628 ± 0.039} & \textbf{0.186 ± 0.019} & \textbf{0.442 ± 0.046} &         \textbf{ 0.098} &         \textbf{ 1.000 }    \\
\hline
 \textbullet & & \textbullet & \textbullet &    & \textbf{0.599 ± 0.075 }& \textbf{0.211 ± 0.046} & \textbf{0.387 ± 0.086} &        \textbf{     0.027} &    \textbf{     0.193} \\
\hline
\textbullet &  & \textbullet &   & \textbullet  & \textbf{0.617 ± 0.061} & \textbf{0.195 ± 0.025} &\textbf{ 0.423 ± 0.064 }&        \textbf{0.076} &         \textbf{0.760 } \\
\hline
\textbullet & &   & \textbullet & \textbullet & \textbf{0.629 ± 0.073 }& \textbf{0.193 ± 0.028 }& \textbf{0.436 ± 0.086} &          \textbf{0.321} &         \textbf{ 1.000 }    \\
\hline
\textbullet & \textbullet & \textbullet & \textbullet &     &\textbf{ 0.624 ± 0.060} & \textbf{0.196 ± 0.029} & \textbf{0.428 ± 0.069} &            \textbf{ 0.135} &        \textbf{ 0.665} \\
\hline
\textbullet & \textbullet & \textbullet &   & \textbullet &    0.587 ± 0.042 & 0.185 ± 0.021 & 0.403 ± 0.048 &              0.001 &        1.000      \\ 
\hline
\textbullet & \textbullet &   & \textbullet & \textbullet & \textbf{0.608 ± 0.053} & \textbf{0.192 ± 0.027} & \textbf{0.417 ± 0.071} &          \textbf{ 0.021} &      \textbf{  1.000 }     \\  
\hline
\textbullet &  & \textbullet & \textbullet & \textbullet  & \textbf{0.609 ± 0.046} & \textbf{0.194 ± 0.023} & \textbf{0.415 ± 0.043} &           \textbf{ 0.017 }&       \textbf{  0.770 } \\
  \hline
\textbullet &   \textbullet & \textbullet & \textbullet & \textbullet & \textbf{0.617 ± 0.058} & \textbf{0.183 ± 0.025} & \textbf{0.434 ± 0.056} &         \textbf{0.065} &       \textbf{  1.000}     \\
\hline 
\end{tabular}%
}
\caption[Combination of Omic Scenarios]{Combination of clinical data with pmic Scenarios for BRCA-SA with a single risk. The \( p \)-values that exceed 0.01 are in bold. The one with the best CI-IBS is shaded in a different colour. CI-p HB and IBS-p HB represent the \( p \)-values corrected with HB for the C-Index and IBS, respectively.}
\label{omicbrca}
\end{table} 
 
In Figure \ref{wsilgg}, we show the evaluation of different numbers of patches for BRCA-CR using clinical data. The \( p \)-values greater than 0.01 are shown in bold. The configuration with the best CI-IBS (defined as the average of the C-Indices across all risks minus the average IBS) is highlighted. CI-p HB and IBS-p HB represent the 
 \( p \)-values for the C-Index and IBS, respectively. In this case, the selected configuration is the one using 10 patches.

\begin{table}[ht]
\centering
\resizebox{0.7\textwidth}{!}{ 
\begin{tabular}{|c|c|c|c|c|c|c|}
\rowcolor[gray]{1}
\hline
 \textbf{N Patches}     &   \textbf{Risk} & \textbf{C-index}     & \textbf{IBS}           &   \textbf{AVG CI-IBS }&  \textbf{ p-CI HB} &    \textbf{p-IBS HB}  \\
\hline
  \textbf{1} & \textbf{1} & \textbf{0.583 ± 0.034} & \textbf{0.231 ± 0.051 }&       \textbf{ 0.369} &           \textbf{0.011} &        \textbf{ 1.000 }    \\
\hline
 \textbf{ 1} &\textbf{ 2}  &\textbf{ 0.622 ± 0.042} & \textbf{0.236 ± 0.057} &       \textbf{ 0.369} &           \textbf{1.000 }     &    \textbf{   0.915 }  \\
\hline
  \textbf{5} &  \textbf{1} & \textbf{0.607 ± 0.049 }& \textbf{0.243 ± 0.049} &       \textbf{ 0.387 }&         \textbf{0.891} &        \textbf{ 0.616} \\
  \hline
  \textbf{5} & \textbf{ 2 }& \textbf{0.625 ± 0.074} & \textbf{0.215 ± 0.031} &      \textbf{ 0.387} &           \textbf{1.000 }    &        \textbf{ 1.000 }    \\
\hline
\rowcolor[HTML]{D8A9C4}\textbf{ 10} &  \textbf{1} & \textbf{0.627 ± 0.036} & \textbf{0.217 ± 0.027} &    \textbf{   0.400}   &      \textbf{  1.000  }   &        \textbf{ 1.000  }   \\
\hline
\rowcolor[HTML]{D8A9C4} \textbf{10} &  \textbf{2 }& \textbf{0.614 ± 0.054} &\textbf{ 0.224 ± 0.043} &       \textbf{  0.400}   &         \textbf{  1.000   }  &       \textbf{  1.000  }   \\ \hline
 \textbf{15} &  \textbf{1} &\textbf{ 0.601 ± 0.057} & \textbf{0.220 ± 0.030} &      \textbf{ 0.380}  &         \textbf{0.675} &        \textbf{ 1.000}     \\  \hline
 \textbf{15} &  \textbf{2} & \textbf{ 0.612 ± 0.052 }& \textbf{0.233 ± 0.060} &      \textbf{  0.380 } &        \textbf{ 1.000 }   &       \textbf{   1.000 } \\
\hline
\end{tabular}%
}
\caption[Combination of Imaging Modalities]{Evaluation of different numbers of patches for BRCA-CR. The \( p \)-values greater than 0.01 are in bold. The configuration with the best CI-IBS (defined as the average of the C-Indices across all risks minus the average IBS) is highlighted. CI-p HB and IBS-p HB represent the \( p \)-values corrected with HB for the C-Index and IBS, respectively.}
\label{wsilgg}
\end{table}

After completing the hyperparameter optimization for all modalities and the selection of the best combination of omics modalities and the optimal number of image patches, we obtained the final results. These are the outcomes discussed in the Results section (see Sections \ref{resultssa} and \ref{resultscr}). 

\section{Multimedia Materials}
This section contains interactive visualizations and exploratory tools that complement the core results of our study. These materials allow dynamic inspection of survival predictions across models and individual patients, as well as gene-specific queries for personalized genomic insights.
\subsection{Interactive Survival Curves by Model}\label{sece1}
 
An interactive visualization comparing the survival curves of different trained models against the KM estimate.
Users can toggle models on/off via the legend and visually compare model behaviors across different data modalities (clinical, omic, histological). The plot also enables inspection of the estimated survival probability at each time point for every model.
\begin{itemize}
    \item \href{https://albagarridolopezz.github.io/SAMVAE/interactive_plot_lgg_sa.html}{Interactive LGG Survival Plot}
  \item \href{https://albagarridolopezz.github.io/SAMVAE/interactive_plot_brca_sa.html}{Interactive BRCA Survival Plot}
  
\end{itemize}

\subsection{Best vs Worst Prognosis Patients} \label{sece2}

A personalized medicine tool that visualizes survival predictions for the best and worst prognosis patients according to the model. For each patient, the plot shows the mean predicted survival curve with 90\% confidence interval and the clinical variables as hover-over tooltips. This tool helps interpret the survival curves derived from clinical data and demonstrates that it is possible to generate personalized survival curves for each patient, thereby promoting personalized medicine and supporting model explainability:
\begin{itemize}
    \item \href{https://albagarridolopezz.github.io/SAMVAE/best_vs_worst_prognosis_interactive.html}{Best vs Worst Prognosis Patients for LGG}
\end{itemize}

Additionally, a Gene CNV Query Notebook is provided. This notebook enables users to check the CNV status of frequently mutated genes for each patient. Beyond visualizing survival curves, users can verify the presence and alteration of specific genes known to be cancer risk indicators on a per-patient basis.

\section{Clinical Explainability}
\begin{figure}[ht]
\centering
\includegraphics[width=0.8\textwidth]{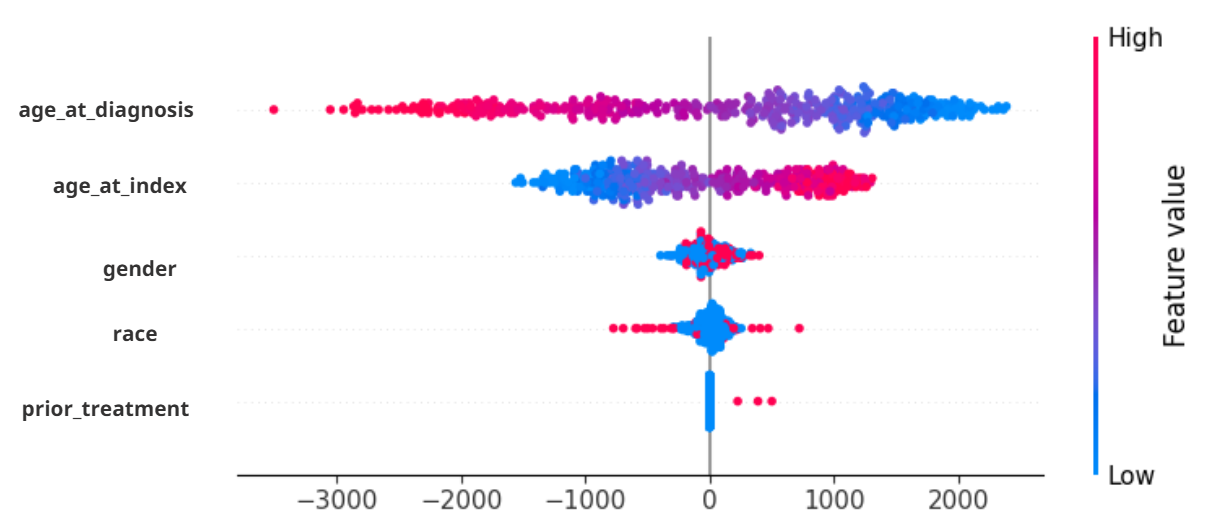}
\caption{Beeswarm plot of SHAP values for LGG-SA. }
\label{xai2}
\end{figure}

\begin{figure}[ht]
\centering
\includegraphics[width=0.8\textwidth]{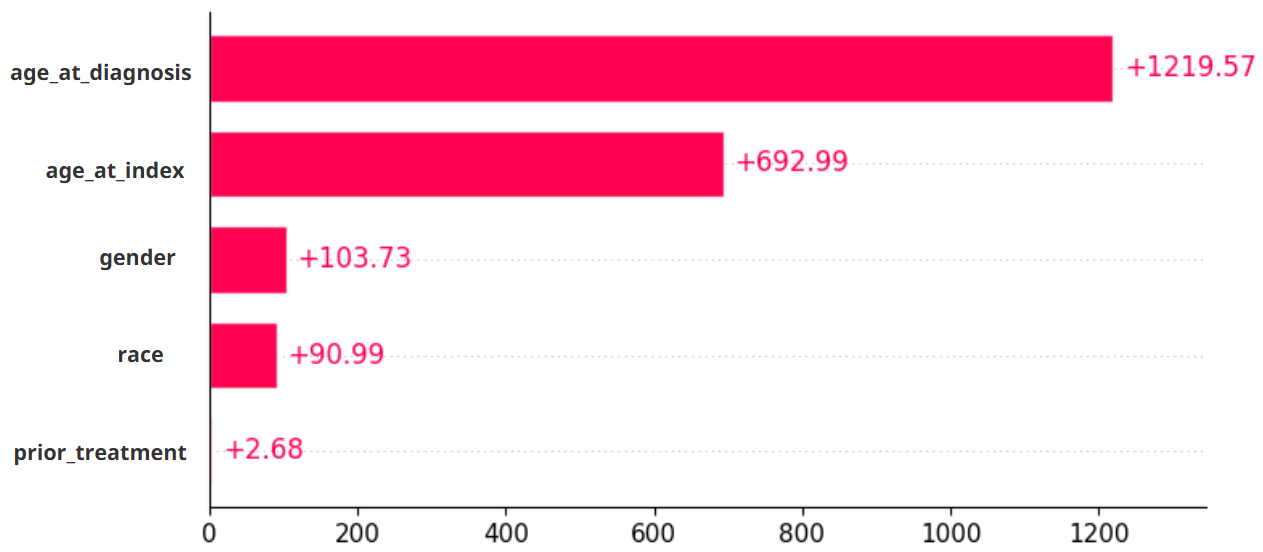}
\caption{Mean Absolute SHAP Values for LGG-SA Clinical Features. }
\label{xai1}
\end{figure}

To improve interpretability, SHAP (SHapley Additive exPlanations)  \cite{bib36} values were computed for the clinical modality. These values quantify the contribution of each clinical feature to individual survival predictions. Beeswarm and bar plots (Figures \ref{xai1} and \ref{xai2}) were generated to visualize both the global importance and local effects of features.

In Figures \ref{xai1} and \ref{xai2}, we can see that age-related features play a dominant role in survival prediction. The model consistently assigns higher importance to “age at diagnosis” and “age at index”, reaffirming that patient age is a key factor in determining prognosis, as previously discussed in \cite{bib23}.

\end{appendices}
 
%Bibliography
\bibliographystyle{ieeetr}
\bibliography{references}

\end{document}